\documentclass[preprint,12pt]{elsarticle}

\usepackage{amssymb}
\usepackage{amsmath}
\usepackage{booktabs}
\usepackage{graphicx}
\usepackage{multirow}
\usepackage{adjustbox}
\usepackage{hyperref}
\usepackage{algorithm}
\usepackage{algorithmic}
\usepackage{xcolor} 
\usepackage[percent]{overpic}
\usepackage{rotating}
\usepackage{float}
\usepackage{tikz}
\usepackage{subcaption}

\author{Hyeonseo Lee}
\author{Juhyun Park}
\author{Jihyong Oh}
\author{Chanho Eom\corref{cor1}} 

\cortext[cor1]{Corresponding author \\
E-mail address: \href{mailto:hswith@cau.ac.kr}{hswith@cau.ac.kr}, \href{mailto:hswith@cau.ac.kr}{juhyunpark@cau.ac.kr}, \href{mailto:jihyongoh@cau.ac.kr}{jihyongoh@cau.ac.kr}, \href{mailto:chanho.eom@cau.ac.kr}{cheom@cau.ac.kr}.}

\address{Graduate School of Advanced Imaging Science, Multimedia \& Film, Chung-Ang University, Seoul, 06974, Korea}

\begin{document}
\sloppy
\begin{frontmatter}

\title{Domain Generalization for Person Re-identification: A Survey Towards Domain-Agnostic Person Matching}


\begin{abstract}
Person Re-identification (ReID) aims to retrieve images of the same individual captured across non-overlapping camera views, making it a critical component of intelligent surveillance systems. Traditional ReID methods assume that the training and test domains share similar characteristics and primarily focus on learning discriminative features within a given domain. However, they often fail to generalize to unseen domains due to domain shifts caused by variations in viewpoint, background, and lighting conditions. To address this issue, Domain-Adaptive ReID (DA-ReID) methods have been proposed. These approaches incorporate unlabeled target domain data during training and improve performance by aligning feature distributions between source and target domains. However, their reliance on access to target domain data limits their scalability and makes them less suitable for real-world deployments, where such data may not be available in advance. Domain-Generalizable ReID (DG-ReID) tackles a more realistic and challenging setting by aiming to learn domain-invariant features without relying on any target domain data. Recent methods have explored various strategies to enhance generalization across diverse environments, but the field remains relatively underexplored. In this paper, we present a comprehensive survey of DG-ReID. We first review the architectural components of DG-ReID including the overall setting, commonly used backbone networks and multi-source input configurations. Then, we categorize and analyze domain generalization modules that explicitly aim to learn domain-invariant and identity-discriminative representations. To examine the broader applicability of these techniques, we further conduct a case study on a related task that also involves distribution shifts. Finally, we discuss recent trends, open challenges, and promising directions for future research in DG-ReID. To the best of our knowledge, this is the first systematic survey dedicated to DG-ReID. 
A curated list of related resources and papers is also available at: \href{https://github.com/PerceptualAI-Lab/Awesome-Domain-Generalizable-Person-Re-ID}{https://github.com/PerceptualAI-Lab/Awesome-Domain-Generalizable-Person-Re-ID}
\end{abstract}


\end{frontmatter}

\section{Introduction}
\label{sec1}
\begin{figure}[t!]
  \centering
  \begin{tikzpicture}
    \node[inner sep=0pt] (img) at (0,0) {\includegraphics[width=\linewidth]{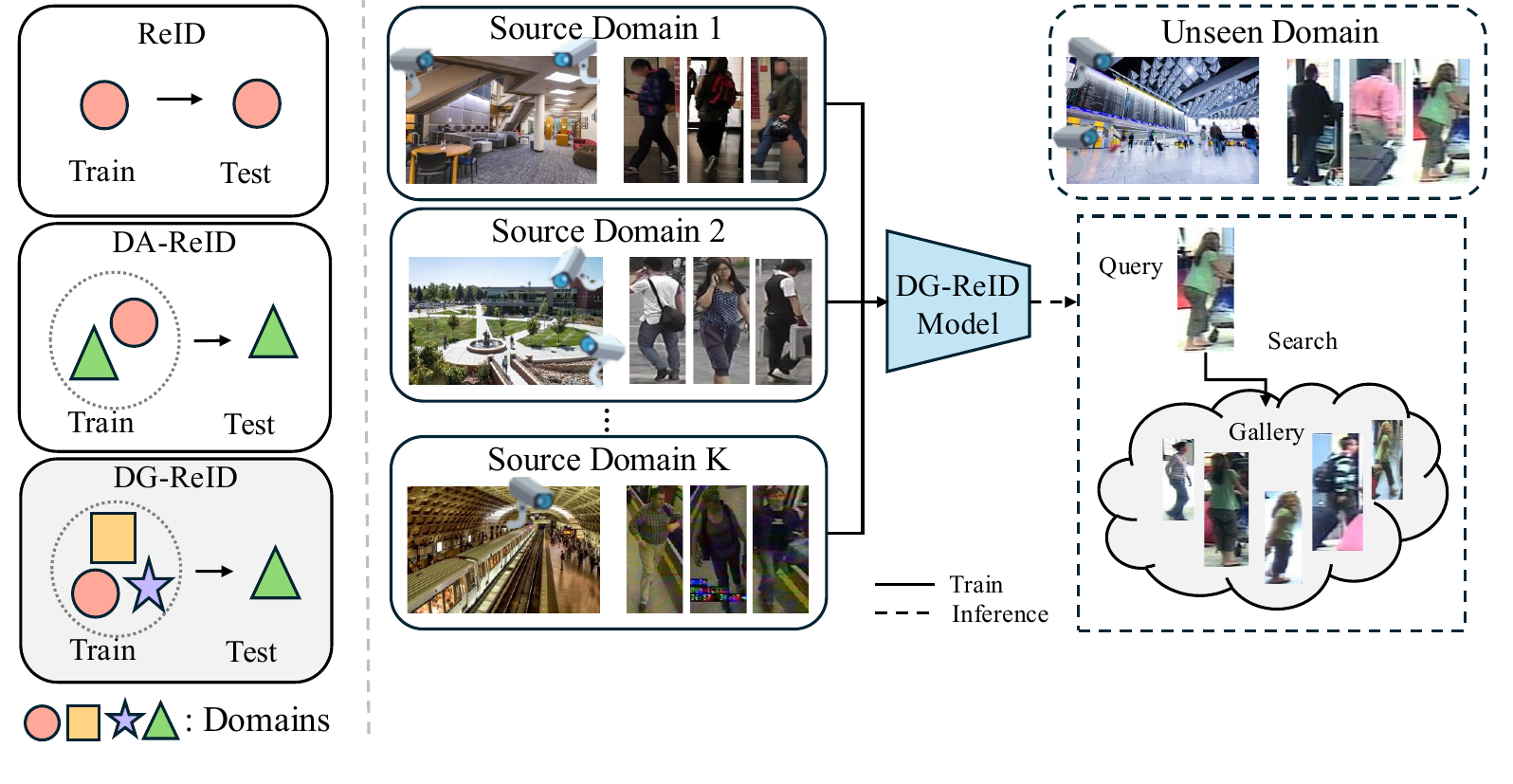}};
    \node[anchor=north west] at ([xshift=40pt,yshift=-200pt]img.north west) {\footnotesize(a)};
    \node[anchor=north east] at ([xshift=-140pt,yshift=-200pt]img.north east) {\footnotesize(b)};
  \end{tikzpicture}
  \vspace{-15pt}
  \caption{Train-test settings in Person ReID and the typical framework of DG-ReID. (a) Comparison of different train-test settings in Person ReID. In the Traditional ReID setting, a model is trained and tested within the same domain, whereas in the DA-ReID setting, a model is trained with source data while adapting to a specific target domain. In the DG-ReID setting, a model is trained on multiple source domains to enhance generalization to unseen domains. (b) In the typical framework of DG-ReID, the model is trained on $K$ source domains (solid arrow) to learn domain-invariant features and is tested on an unseen domain (dashed arrow) by retrieving matching gallery images for a query.}
  \vspace{-10pt}
  \label{fig:framework}
\end{figure}

Person Re-identification (ReID) aims to retrieve images of the same individual across non-overlapping camera views. ReID is crucial for surveillance system, particularly in locating missing persons and tracking criminals~\cite{zheng2015scalable, zheng2017gan, li2014deepreid,yu2024smagnet}. Traditional ReID methods~\cite{zheng2015scalable, zheng2017gan, li2014deepreid, wei2018gan, sun2018beyond, mgn2018, eom2019disentangled, zhu2020identity, zhao2017spindlenet, eom2022disentangled} assume that training and test domains share similar characteristics (Fig. \hyperref[fig:framework]{1} (a) top), and primarily focus on extracting discriminative features within a given domain. However, in real-world scenarios, this assumption often fails as domain shifts, such as variations in camera viewpoints, backgrounds, and lighting conditions, create significant distributional discrepancies, hindering generalization to unseen domains.
To address this problem, Domain-Adaptive ReID (DA-ReID) methods have been proposed. These methods improve performance by using target domain data during training (Fig.~\hyperref[fig:framework]{1} (a) middle), using images and labels~\cite{panda2017unsupervised} or only labels~\cite{chen2020deep, lee2023cameradriven}. However, while DA-ReID methods outperform traditional ReID methods on a specific target domain, they often learn domain-specific features that overfit to the target domain, which limits their generalization to unseen domains. To overcome this limitation, recent research has focused on Domain-Generalizable ReID (DG-ReID) methods~(Fig. \hyperref[fig:framework]{1} (a) bottom). Unlike DA-ReID, DG-ReID methods do not rely on target domain data but instead aim to achieve stable performance across various domains. As illustrated in Fig. \hyperref[fig:framework]{1} (b), the DG-ReID model is trained on $K$ distinct source domains, each with different backgrounds and camera viewpoints. Once trained,  the model is evaluated on an unseen domain to retrieve matching images of the same identity from the gallery. This approach demonstrates robust generalization to environments unseen during training. However, despite the strong potential of DG-ReID methods for real-world scenarios, research in this area remains relatively underexplored.

\begin{figure}[t!]
    \centering
    \includegraphics[width=\linewidth]{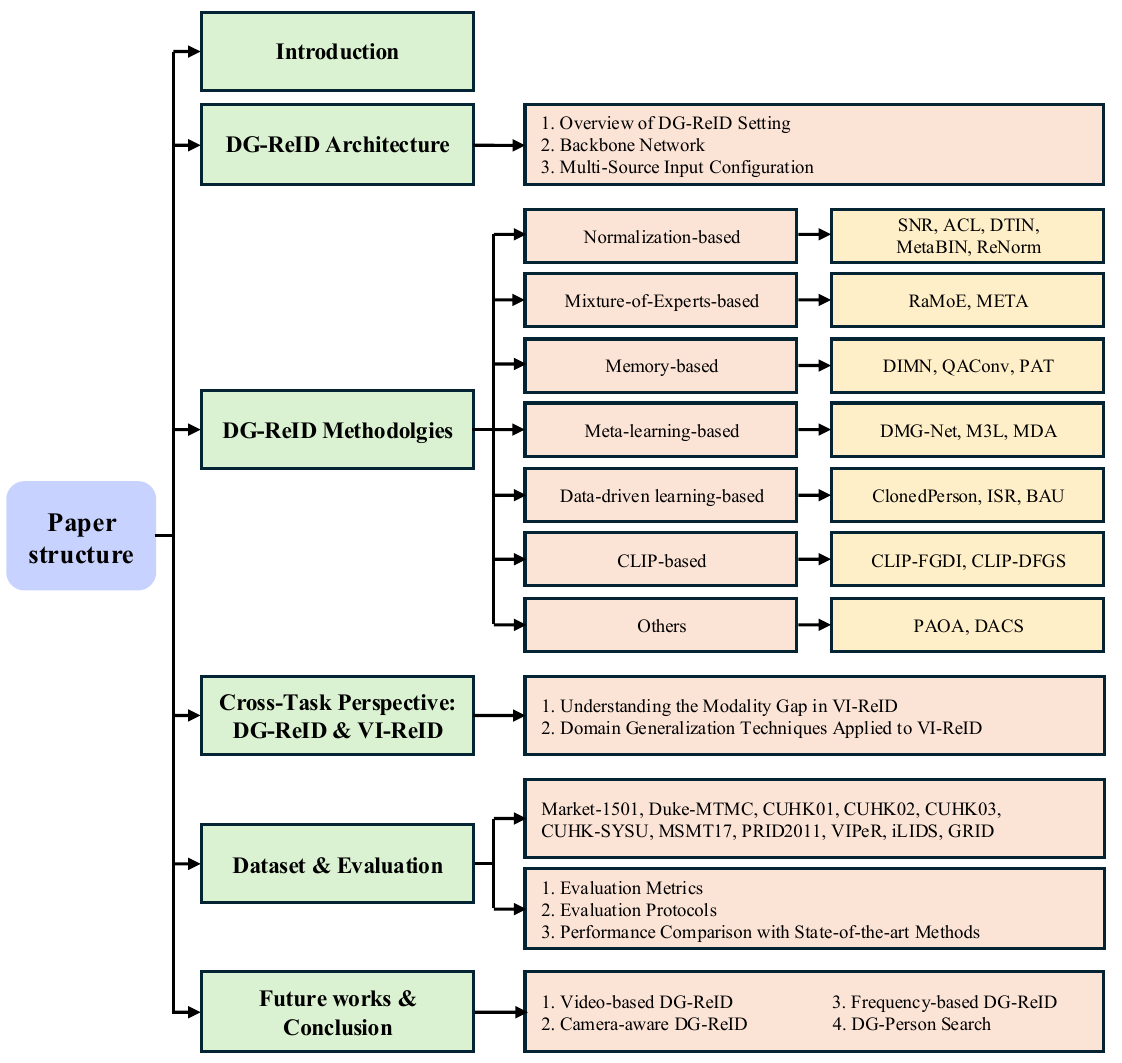}
    \vspace{-15pt}
    \caption{Taxonomy of Domain Generalizable person re-identification (DG-ReID).}
    \label{fig:tax}
\end{figure}

To facilitate future advancements, we present a comprehensive survey\footnote{We collected DG-ReID methods published between 2018 and 2025 in peer-reviewed venues, primarily focusing on approaches with methodological contributions to generalization. The literature collection involved keyword-based searches (\textit{e.g.}, ``domain generalizable re-identification'') on academic databases such as Google Scholar and IEEE Xplore, among other sources.} that systematically categorizes and analyzes existing methods, offering a structured overview of the main pipeline and recent advancements in DG-ReID. Our survey aims to help researchers quickly grasp key directions and emerging trends in the field. To this end, we focus on two central aspects of DG-ReID. Firstly, we investigate what fundamentally distinguishes DG-ReID from traditional Re-ID, particularly in terms of system-level assumptions and training constraints. Unlike conventional settings, DG-ReID operates without access to target domain data, necessitating architectural and sampling designs that promote robustness under unseen conditions. Secondly, we explore how recent methods achieve such generalization by introducing dedicated DG modules. These modules--ranging from normalization strategies to expert systems and language-guided priors--form the core mechanisms for overcoming domain bias. These key aspects shape the foundation of this survey, allowing us to consolidate and interpret the field through a principled and task-aware perspective. These two analytical axes serve as the basis for the structure of our survey. In addition to DG-ReID, we also explore how domain generalization techniques extend to other related tasks. Specifically, we include a case study on \textit{Visible-Infrared Re-ID (VI-ReID)}, which involves bridging the modality gap between visible and infrared images. Although traditionally studied as a separate field, VI-ReID shares conceptual similarities with DG-ReID in addressing distribution shifts, offering valuable cross-task insights.

To the best of our knowledge, this is the first comprehensive survey dedicated to Domain Generalization in Person Re-Identification (DG-ReID), system-level training strategies, covering core methodologies, and open research challenges. The remainder of this paper is organized as follows. Section~\hyperref[sec2]{2} introduces the architectural components of DG-ReID, including problem definition, backbone networks, and multi-source input configurations. Section~\hyperref[sec3]{3} presents a taxonomy of existing methodologies, categorized into normalization-based, mixture-of-experts-based, memory-based, meta-learning-based, data-driven learning-based, CLIP-based, and other approaches. Section~\hyperref[sec4]{4} explores cross-task insights through a case study on VI-ReID. Section~\hyperref[sec5]{5} reviews commonly used datasets and evaluation protocols. Finally, Section~\hyperref[sec6]{6} outlines future directions. An overview of this paper structure is illustrated in Fig.~\hyperref[fig:tax]{2}.

\section{DG-ReID Architecture}
\label{sec2}
To provide a structured overview of Domain Generalizable Person Re-identification (DG-ReID), this section outlines the standard architectural components commonly used across methods. We highlight widely adopted backbone networks and describe how multiple source domains are configured and sampled during training. Specifically, Section~\hyperref[subsec2-1]{2.1} introduces the DG-ReID problem setting, Section~\hyperref[subsec2-2]{2.2} reviews representative backbone architectures, and Section~\hyperref[subsec2-3]{2.3} examines multi-source input configuration strategies.

\subsection{Overview of DG-ReID Setting}
\label{subsec2-1}
Domain Generalizable Person Re-identification (DG-ReID) aims to identify individuals in unseen domains without accessing target domain data during training. Each source domain contains mutually exclusive identity labels and domain-specific biases (\textit{e.g.}, background, lighting, viewpoint). Due to the non-overlapping identities across domains, direct identity matching is infeasible, making it essential to learn features that are both domain-invariant and identity-discriminative. Formally, let $\mathcal{D}_{S} := \{ \mathcal{D}_{k} \}_{k=1}^{K}$ denote a set of $K$ source domains, where each domain is defined as $\mathcal{D}_k := \{ (x_{i}, y_{i}) \mid x_{i} \in \mathcal{X}, y_{i} \in \mathcal{Y}_{k} \}_{i=1}^{N_k}$. The identity label spaces are disjoint, \textit{i.e.}, $\mathcal{Y}_{i} \cap \mathcal{Y}_{j} = \emptyset$ for all $i \ne j$, characterizing DG-ReID as a \textit{heterogeneous} domain generalization problem. A feature extractor $f_{e_\theta}$ is trained on the union of source domains to learn identity-relevant features. During inference, the model is evaluated on a target domain $\mathcal{D}_T$ that was not seen during training and contains entirely new identities.

\subsection{Backbone Network}
\label{subsec2-2}
\paragraph{ResNet}
ResNet-50~\cite{he2016deep}, pre-trained on ImageNet, is the most commonly used backbone in DG-ReID. Its wide adoption stems from both practical and methodological reasons: (1) it enables fair comparison of DG modules under a shared architecture, and (2) due to the relatively uniform structure of ReID datasets--centered subjects and fixed resolution--convolutional inductive bias of ResNet remains effective even in DG scenarios. A key trend in recent works is to modify the normalization layers within ResNet to improve generalization. For example, SNR~\cite{jin2020style} augments the backbone by inserting Instance Normalization (IN) layers to suppress style-related variations. MetaBIN~\cite{choi2021metabin}, BAU~\cite{cho2024alignment}, and ISR~\cite{dou2023identity} replace or refine the original Batch Normalization (BN) layers with hybrid or adaptive schemes. Notably, MetaBIN computes a learnable combination of BN and IN outputs during training, enabling the model to adaptively balance domain-specific and domain-invariant components. These methods retain the overall ResNet structure and focus on enhancing normalization behavior through lightweight, learnable modules.
In contrast, IBN-Net50~\cite{pan2018two} restructures early convolutional layers by splitting feature channels, applying IN to one part and BN to the other. This architectural design contrasts with methods that merely insert or mix IN and BN within standard normalization layers. Specifically, a fixed portion of feature channels in shallow layers is normalized with IN to reduce style variance (\textit{e.g.}, illumination or background), while deeper layers retain BN to preserve discriminative power. This architectural design is not adjusted during training but has shown strong generalization in DG tasks, as demonstrated in M3L~\cite{zhao2021memory}. DTIN-Net~\cite{jiao2022dtin} introduces dynamic convolution operations into the backbone to enhance generalization under domain shift. It features a dynamic control path that modulates feature representations by aggregating context from multiple stages of the network, allowing the receptive field to adapt based on input characteristics. While still grounded in a ResNet-like architecture, DTIN-Net departs from conventional CNNs by embedding flexibility into the convolutional process itself, offering a unified framework for learning domain-robust identity features. Further architectural details are provided in Section~\hyperref[subsec3-1]{3.1}. These approaches highlight how both modular normalization designs and structural revisions of ResNet play a key role in extracting domain-invariant features for DG-ReID.

\paragraph{MobileNet}
Several DG-ReID methods adopt MobileNet-based architectures to achieve a favorable trade-off between efficiency and performance. Unlike deeper models such as ResNet-50, MobileNetV2~\cite{sandler2018mobilenetv2} employs depthwise separable convolutions, significantly reducing the number of parameters while maintaining expressive power. DIMN~\cite{song2019dimn} utilizes MobileNetV2 as the encoding subnetwork to enable low-cost deployment of DG models. MetaBIN~\cite{choi2021metabin} further explores a wider variant of MobileNetV2 (width multiplier of 1.4) to balance representation capacity and computational efficiency. The structural simplicity of MobileNet offers advantages beyond computational efficiency, making it meaningful even from a domain generalization perspective. First, its limited model capacity helps prevent overfitting to complex style or background patterns in the source domain, thereby encouraging the learning of more general appearance-based identity features. Second, by avoiding the deep hierarchical inductive biases embedded in large CNNs, MobileNet may retain more stable feature representations under unseen domain conditions. Lastly, due to its lightweight architecture, MobileNet enables seamless integration with various DG modules such as normalization, meta-learning, and memory-guided strategies. Its moderate representational strength also allows the effects of each module to be more clearly observed during experimentation. Thus, MobileNet is not merely a lightweight alternative but a structurally interpretable and experimentally controllable backbone, making it a valuable choice for DG-ReID research.

\paragraph{ViT}
PAT~\cite{ni2023pat} adopts ViT-B/16~\cite{dosovitskiy2020vit}, a Vision Transformer pre-trained on ImageNet with a patch size of $16 \times 16$, as its backbone. Unlike convolutional networks that rely on local receptive fields, ViT exhibits a structural distinction by modeling global context across all input tokens through self-attention. This design is advantageous for capturing long-range dependencies, such as relationships between distant body parts, and for building robust representations under background clutter or pose variation. Moreover, because ViT lacks strong inductive biases--such as locality and translation invariance--it is relatively less susceptible to overfitting to domain-specific styles. When adequately trained, ViT holds the potential to learn more generalizable visual representations. However, this lack of inductive bias can also lead to instability in low-data or source-biased training scenarios. To address this limitation, PAT introduces a mechanism that encourages the model to learn local similarity among different body parts, effectively injecting a form of localized inductive bias into ViT. This helps guide the learning of representations that are both domain-invariant and discriminative. In doing so, PAT leverages structural flexibility and global expressiveness of ViT to enable identity-consistent representation learning under domain shifts, demonstrating the potential of ViT as a strong backbone for DG-ReID. Beyond pure ViT-based models, ViT-B/16 also serves as the image encoder in vision-language frameworks such as CLIP~\cite{radford2021clip}. In CLIP, ViT is pre-trained jointly with a text encoder to project visual features into a semantically aligned embedding space. CLIP-based methods, including CLIP-FGDI~\cite{zhao2025clipfgdi} and CLIP-DFGS~\cite{zhao2024clipdfgs}, utilize this ViT encoder without structural modification, while incorporating domain-aware mechanisms tailored to domain generalization. CLIP-FGDI generates domain-specific prompts that intentionally induce inter-domain confusion, while simultaneously encouraging attention to foreground cues via domain-invariant prompts. This allows ViT to extract more robust and discriminative features. CLIP-DFGS, on the other hand, refines descriptive identity texts and guides the ViT encoder to form image embeddings that are closer for the same identity and farther for different ones. These approaches extend the expressive power of ViT not by modifying its architecture, but by leveraging text-guided alignment and domain-specific modulation. As a result, ViT functions as a central backbone for cross-domain identity matching, offering strong global context modeling and semantically structured embedding capabilities.

\subsection{Multi-Source Input Configuration}
\label{subsec2-3} 
A common strategy in DG-ReID is to utilize multiple source datasets to enhance domain diversity during training. Most approaches~\cite{jin2020style, zhang2022adaptive, nie2025normalization, choi2021metabin, xu2022mimic, dou2023identity, ni2023pat, cho2024alignment, li2024paoa} merge all source datasets into a single training pool, where identity labels are offset to avoid collision across domains. This unified setup assumes that identities do not overlap across datasets, and thus label indices are incrementally shifted for each domain. Alternatively, a few methods adopt a domain-specific training configuration~\cite{dai2021ramoe}, where each source dataset is processed through a separate network or adapter, and model parameters are later aggregated through meta-learning or mixture-of-experts frameworks. Regardless of the merging strategy, both approaches commonly rely on the identity-based random sampling mechanism~\cite{hermans2017defense} to construct mini-batches. This sampler selects $P$ identities and $K$ images per identity, without considering domain balance. The dominance of identity-level learning explains why domain balancing is often neglected, especially when the number of source domains is relatively small (typically 3–4). Empirical evidence suggests that domain imbalance in such settings may not significantly degrade generalization performance. However, domain-aware labels are still retained in many works~\cite{zhang2022adaptive, choi2021metabin}, enabling auxiliary learning signals such as domain adversarial training~\cite{song2019dimn}, domain-specific feature disentanglement~\cite{ni2023pat}, or dynamic expert selection~\cite{dai2021ramoe}. These methods leverage domain labels as soft structural priors, even when domains are merged. Overall, the multi-source input strategy in DG-ReID reflects a trade-off: merging domains maximizes data efficiency and representation diversity, while preserving domain-specific structures enables better modeling of inter-domain variation.

\section{DG-ReID Methodologies}
\label{sec3}
This section introduces domain generalization (DG) methodologies for person re-identification from a module-centric perspective. We categorize existing approaches into eight methodological aspects, including normalization-based, mixture-of-experts-based, memory-based, meta-learning-based, data-driven learning-based, CLIP-based, and other emerging methods. Each category reflects a distinct strategy for mitigating domain shift and enhancing cross-domain generalization in DG-ReID.

\subsection{Normalization-based}
\label{subsec3-1}
Normalization plays a crucial role in enhancing generalization by regulating feature distributions and reducing domain discrepancies~\cite{ulyanov2016instancenorm, ioffe2015batchnorm}. Among the most widely used techniques are BN~\cite{ioffe2015batchnorm} and IN~\cite{ulyanov2016instancenorm}, provide complementary advantages in the context of domain generalization. BN normalizes features using batch-level statistics, capturing shared characteristics within a batch and often preserving domain-specific cues such as lighting or background~\cite{li2016batchnormda, chang2019dsbn}. This property is commonly exploited to model domain-specific distributions in DG-ReID. In contrast, IN normalizes each instance independently, effectively removing style-related variations such as color and texture~\cite{huang2017adain, nam2019styleagnostic}. This enables grouping of same-identity samples across domains, making IN well-suited for extracting domain-invariant features. While BN and IN are individually effective, they exhibit complementary strengths and limitations. BN captures domain-specific styles due to statistics computed from same-domain batches, leading to reliance on domain-dependent cues and poor generalization. This stands in opposition to the central objective of DG, which is to learn features invariant to domain-specific appearance shifts.~\cite{huang2017adain}. IN enhances domain invariance but sometimes at the cost of losing identity-distinctive cues. As such, several DG-ReID methods have explored strategies to balance these trade-offs by combining BN and IN within the same network. Normalization-based DG-ReID methods are typically divided into fixed (\textit{e.g.}, IBN~\cite{pan2018ibn}) and learnable (\textit{e.g.}, BIN~\cite{choi2021metabin}) combinations of BN and IN. While IBN statically splits channels, BIN assigns dynamic weights during training. These hybrid strategies aim to balance domain invariance and identity discrimination for improved generalization.

\begin{figure}[t!]
    \centering
    \includegraphics[width=\linewidth]{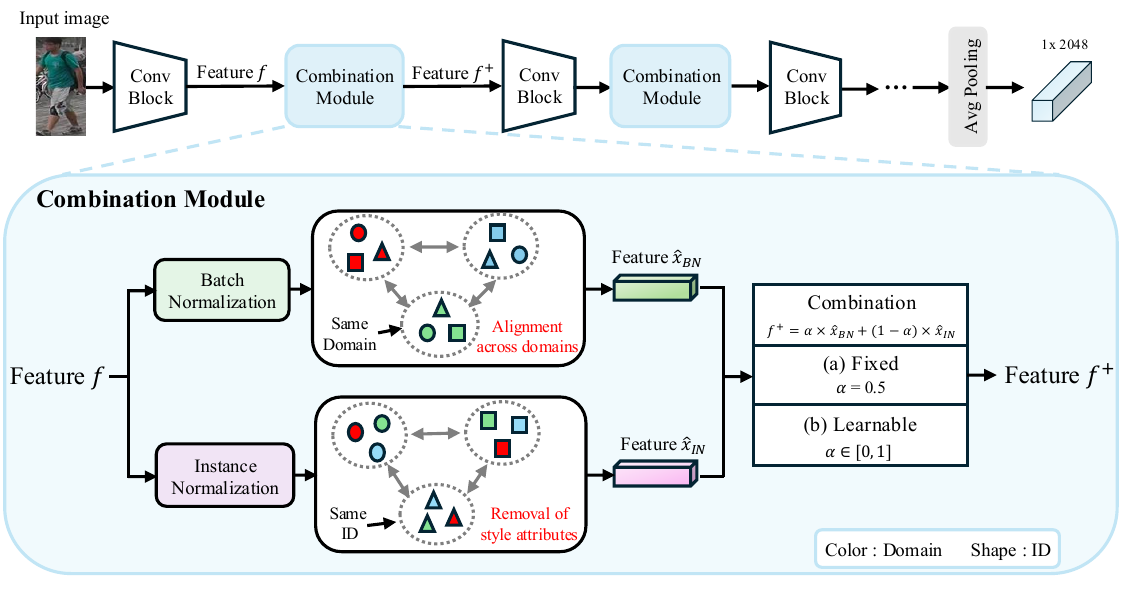}
    \vspace{-15pt}
    \caption{Visualization of Normalization Effects on Feature Distribution. BN}~\cite{ioffe2015batchnorm} groups samples by domain, while IN~\cite{ulyanov2016instancenorm} clusters them by ID. Fixed combination methods~\cite{jin2020style, zhang2022adaptive, xu2022mimic, nie2025normalization} use a preset ratio $\alpha=0.5$, whereas learnable methods~\cite{choi2021metabin, nie2025normalization} adjust $\alpha \in [0, 1]$ dynamically.
    \vspace{-10pt}
    \label{fig:norm}
\end{figure}

\paragraph{Fixed Combination} 
Fixed combination methods~\cite{jin2020style, zhang2022adaptive, xu2022mimic, nie2025normalization} integrate BN~\cite{ioffe2015batchnorm} and IN~\cite{ulyanov2016instancenorm} using predefined strategies without learnable adaptation. These approaches are typically simple in design, involving either complete replacement of BN layers with IN or fixed splits between the two types of normalization~\cite{jin2020style}. For instance, Style Normalization and Restitution (SNR)~\cite{jin2020style} introduces IN into each bottleneck layer of shallow networks to suppress style variations arising from camera views and backgrounds. While this promotes domain-invariant feature learning, it often results in loss of identity-discriminative cues, leading to degraded performance in terms of fine-grained classification. To address this, Adaptive Cross-domain Learning (ACL)~\cite{zhang2022adaptive} employs instance-batch normalization (IBN)~\cite{pan2018ibn} as a more balanced solution. As shown in Fig.~\hyperref[fig:norm]{3} (a), illustrates a generalized hybrid normalization module that combines BN and IN to balance domain-specific discriminability and domain-invariant robustness. As shown, the feature is first normalized by both BN and IN branches, and the results are fused into a single representation using a fixed ratio (\textit{e.g.}, $\alpha = 0.5$). This hybrid strategy aims to preserve both domain-specific and domain-invariant components simultaneously. Subsequent method such as META~\cite{xu2022mimic} builds upon this fixed structure. However, despite its simplicity and stability, the inflexible design of IBN makes it sensitive to shifts in domain distributions, limiting its ability to adapt to unseen domains~\cite{choi2021metabin}.

\paragraph{Learnable Combination} 
Learnable combination methods~\cite{choi2021metabin, nie2025normalization} dynamically balance the use of BN and IN through trainable parameters, allowing the model to adapt normalization behavior based on the input distribution. These approaches are designed to overcome the limitations of fixed structures by learning the optimal trade-off between generalization and discriminative power during training. DTIN~\cite{jiao2022dtin} proposes Dynamically Transformed Instance Normalization to address the limitations of conventional IN. Specifically, a dynamic parameter generator produces convolutional weights and biases conditioned on the input feature map, enabling query-dependent transformations. This allows the model not only to remove style-related variations, but also to adaptively reshape the feature space according to the characteristics of individual instances and domains, thereby achieving both domain invariance and identity discriminability. MetaBIN~\cite{choi2021metabin}, a representative method in this category, proposes to adaptively combine BN and IN using a channel-wise learnable weight $\alpha$:

\begin{equation}
\label{eq1}
\mathbf{y} = \alpha \left( \gamma_{BN} \cdot \hat{{x}}_{BN} + \beta_{BN} \right) + (1 - \alpha) \left( \gamma_{IN} \cdot \hat{{x}}_{IN} + \beta_{IN} \right),
\end{equation}
here, $\hat{x}$ refers to the normalized features, and $\gamma$ and $\beta$ are affine parameters specific to each normalization type. As shown in Fig.~\hyperref[fig:norm]{3} (b), the normalization module continuously updates the parameter $\alpha \in [0, 1]$, defined per channel, controls the relative contribution of BN and IN, allowing the model to flexibly respond to the statistical characteristics of the input. This learnable formulation enables normalization to dynamically emphasize domain-specific or invariant features depending on the training context. Building on this idea, ReNorm~\cite{nie2025normalization} introduces a more sophisticated mechanism that combines two novel normalization types: Remix Normalization and Emulation Normalization. Remix Normalization selectively mixes statistics across domains to enhance intra-source discrimination, while Emulation Normalization simulates unseen domain conditions during training. ReNorm uses a dual forward pass mechanism, where the same input passes through both normalization paths, and their outputs are jointly optimized. This process effectively bridges the training-testing domain gap and improves robustness.

Fixed and learnable combination methods offer contrasting strengths in terms of simplicity, adaptability, and computational cost. Fixed methods statically divide normalization channels, maintaining a lightweight design and enabling fast inference. However, their rigid configuration lacks the flexibility to respond to diverse domain statistics or sample-specific variations, which can limit generalization under domain shift. In contrast, learnable methods employ trainable parameters to dynamically adjust the balance between BN and IN, enabling fine-grained control over normalization behavior. This adaptability enhances performance in unseen domains but introduces higher computational complexity and potential training instability. As normalization alone cannot fully eliminate domain bias, recent research trends increasingly treat fixed combination methods not as standalone solutions but as plug-in modules within larger architectures. Meanwhile, learnable methods continue to evolve as central components for building domain-aware normalization schemes that respond to training context and data characteristics.

\subsection{Mixture-of-Experts-based}
\label{subsec3-2}
The mixture-of-experts (MoE) framework~\cite{jacobs1991adaptive, jordan1994hierarchical} is a modular learning paradigm in which a collection of specialized subnetworks, called ``experts'' collaboratively solve a task by partitioning the problem space. Each expert is trained to handle a particular subset of the input distribution, and a gating mechanism determines the contribution of each expert during inference. Originally developed for divide-and-conquer strategies in machine learning, MoE has gained recent interest in various areas such as multi-task learning, domain adaptation, and generalization~\cite{zhong2022metadmoe, wang2023moelearner}. In the context of DG-ReID, MoE-based methods aim to exploit the diversity of source domains by assigning different experts to model different domain-specific characteristics. By explicitly modeling inter-domain variations through multiple expert branches, the MoE framework allows the network to learn a richer and more decomposable representation that can generalize better to unseen domains. During inference, the gating mechanism adaptively combines experts based on the input data, implicitly estimating its relevance to the known domains. Recent MoE-based DG-ReID approaches can be broadly categorized into two types: those that maintain independent experts, where each expert is solely responsible for a specific source domain, and those that use shared experts across domains, promoting cross-domain knowledge transfer. In the following section, we review these two categories and analyze their impact on generalization performance in DG-ReID settings.

\begin{figure}[t!]
  \centering
  \begin{tikzpicture}
    \node[inner sep=0pt] (img) at (0,0) {\includegraphics[width=\linewidth]{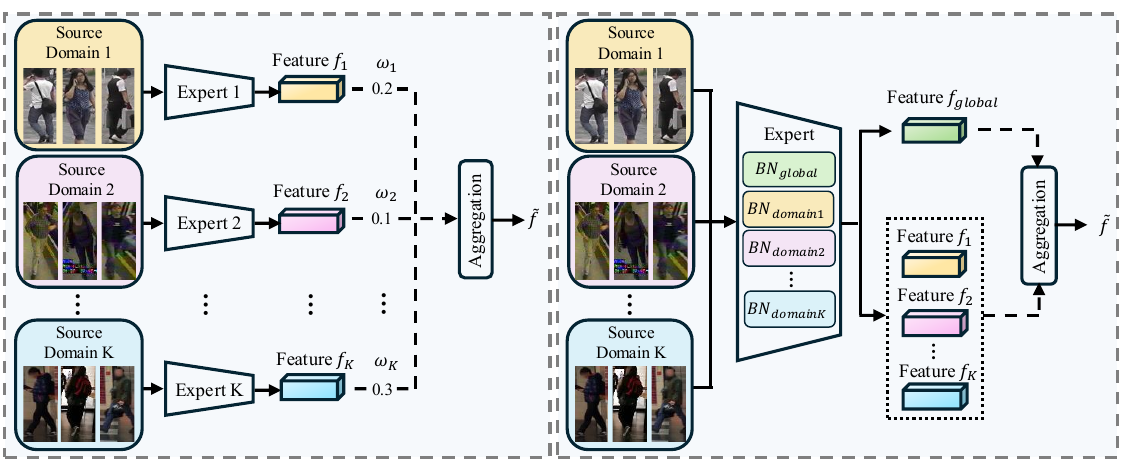}};
    \node[anchor=north west] at ([xshift=90pt,yshift=-160pt]img.north west) {\footnotesize(a)};
    \node[anchor=north east] at ([xshift=-90pt,yshift=-160pt]img.north east) {\footnotesize(b)};
  \end{tikzpicture}
  \vspace{-25pt}
  \caption{Comparison of Independent and Shared Expert Methods. (a) Independent expert methods~\cite{dai2021ramoe} assign separate experts to each source domain and aggregate their outputs into a unified feature $\tilde{f}$. (b) Shared expert methods~\cite{xu2022mimic} use a single expert module with shared parameters across domains, except for BN layers. The model comprises a global branch for general features and an expert branch for domain-specific features, aggregating all outputs.}
  \vspace{-10pt}
  \label{fig:expert_methods}
\end{figure}

\paragraph{Independent experts} 
Relevance-Aware Mixture of Experts (RaMoE)~\cite{dai2021ramoe} introduces a expert branch for each individual source domain. As illustrated in Fig.~\hyperref[fig:expert_methods]{4} (a), each expert is trained exclusively on a specific domain and specializes in capturing domain-specific features and distributional patterns. During inference, the outputs from these domain-specific experts are aggregated using a voting network, which integrates their predictions to form a final decision. To estimate domain relevance, the voting network compares the query feature with class prototypes from each source domain using inner product similarity. The relevance score \( \omega_j \) is computed as the average similarity between the query feature and the class prototypes of the \( j \)-th domain, reflecting how closely the image aligns with that domain’s semantic structure. For each source domain \( j \ne k \), the corresponding expert outputs a domain-specific feature \( f_j \). Using these relevance scores, the final aggregated feature \( \tilde{f} \) is computed by a weighted sum over the expert features:

\begin{equation}
\label{eq3}
\tilde{f} = \sum_{j \ne k} \sigma(\omega_j) \cdot f_j,
\end{equation}
where \( \sigma(\cdot) \) denotes a non-linear normalization function, such as softmax or sigmoid, that maps the relevance scores into the range \([0,1]\). This weighted aggregation enables the model to selectively emphasize features from more relevant source domains during inference. To ensure that the aggregated feature is as discriminative as the original domain-specific feature, a relation alignment loss is introduced. It encourages the voting network to assign accurate relevance scores by comparing the metric similarity between aggregated and domain-specific features. A softmax-based distance function and a binary cross-entropy loss are used to supervise this alignment.

\paragraph{Shared experts} 
Shared expert methods reduce computational redundancy by allowing parameter sharing across domains, while preserving domain-adaptive flexibility. META~\cite{xu2022mimic} explains this strategy by implementing a single expert structure that is reused across all domains, as shown in Fig.~\hyperref[fig:expert_methods]{4} (b). The key innovation lies in the use of dual BN layers: a global BN for capturing domain-invariant representations and expert-specific BN layers for modeling domain-specific statistics. This hybrid normalization enables the model to benefit from shared representations while retaining the capacity to adapt to each domain. By decoupling the domain-specific adaptation from the core network parameters, META~\cite{xu2022mimic} achieves a better trade-off between generalization and efficiency. It significantly reduces the number of parameters compared to fully independent expert models while maintaining competitive performance. Furthermore, the use of shared expert blocks promotes knowledge transfer among domains, which can improve generalization to unseen domains. However, this approach may be less effective when domains are highly dissimilar, as shared parameters may not sufficiently capture the unique characteristics of each domain. Overall, the choice between independent and shared experts reflects a fundamental trade-off in DG-ReID modeling: independent experts offer strong domain-specific specialization at the cost of scalability, while shared experts promote efficiency and knowledge sharing, potentially at the expense of fine-grained domain alignment. As shown in Fig.~\hyperref[fig:expert_methods]{4} (b), the expert network in META~\cite{xu2022mimic} shares all convolutional layers across domains, while separating the normalization paths. Specifically, $BN_{global}$ serves as a common BN layer across all domains to capture domain-invariant features, whereas $BN_{domain}$ layers retain domain-specific statistics for each source domain, enabling the model to preserve domain-dependent characteristics. This design allows the network to simultaneously learn both domain-invariant and domain-specific representations. In addition, a global branch is constructed with IN layers to further enhance the domain-invariant representation. The output features from the expert branch ($f_{domain}$) and the global branch ($f_{global}$) are then combined via a weighted aggregation mechanism. The final feature is obtained by applying a softmax-based weighting over the branches, enabling adaptive fusion of global and domain-specific cues.

Independent expert models assign separate expert branches to each source domain, enabling precise modeling of domain-specific features. However, their scalability is limited as the number of domains increases, leading to high memory and computational costs. In contrast, shared expert models use a single network across all domains, with separate BN layers to retain domain-specific statistics. This design improves efficiency and facilitates knowledge transfer, though it may struggle to capture fine-grained differences when domains are highly diverse. Key challenges in practical applications include the trade-off between scalability and representational power, the difficulty of accurately estimating domain relevance for expert selection, and the ambiguity of domain boundaries in real-world data. Moreover, effectively combining domain-invariant and domain-specific features remains an open problem, requiring careful design of aggregation mechanisms.

\subsection{Memory-based}
\label{subsec3-3}
To improve the generalizability of ReID models under domain shifts, recent studies~\cite{liao2019adaptiveconv, zhong2019em} have introduced memory-based mechanisms that store and utilize feature representations during training or inference. These methods commonly maintain a memory bank of intermediate feature maps, classifier weights, or token representations extracted from the network backbone. By referencing these stored features, the model can adaptively refine current representations or perform more precise comparisons across identities. This design offers two main advantages: (1) it preserves the discriminative power of the learned features by retaining class- or part-level information over time, and (2) it allows for query--adaptive matching, which enhances robustness to intra- and inter-domain variations. Representative approaches such as DIMN~\cite{song2019dimn}, QAConv~\cite{liao2020qa}, and PAT~\cite{ni2023pat} each explore different forms of memory structures and interaction strategies to support discriminative and transferable representation learning.

\paragraph{Contextual memory} 
DIMN~\cite{song2019dimn} formulates the training dataset as a pair of gallery and probe (query) images, sampled at each iteration. Two encoding subnetworks with shared weights are used to extract features from the probe and gallery images, respectively. The probe features are supervised by both classification and triplet losses to enhance identity discrimination, while the gallery features are used to predict classifier weights, which are not used directly for classification but instead stored in a memory bank. The memory bank is realized by a weight matrix $W \in \mathbb{R}^{D \times C}$. In each mini-batch, one image is randomly sampled from each of the $C_b$ identity classes, and used to estimate the corresponding classifier weights $\{\hat{\theta}_{\cdot,j}\}_{j=1}^{C_b}$. These predicted weights are used to update the online memory $\hat{W}$, a copy of $W$, by replacing the corresponding columns:

\begin{equation}
\hat{W}_{\cdot, L(j)} \leftarrow \hat{\theta}_{\cdot, j} \quad \forall j \in [1, 2, \ldots, C_b],
\label{eq:dimn_online_update}
\end{equation}
where $L(j)$ denotes the label index function that maps each mini-batch sample to its column in the memory. The original memory $W$ is then updated using the online memory via a momentum-based moving average:
\begin{equation}
W \leftarrow (1 - \alpha) W + \alpha \hat{W},
\label{eq:dimn_target_update}
\end{equation}
here, the coefficient $\alpha$ controls the update rate and plays a similar role to momentum, allowing the memory to evolve gradually. By maintaining a memory of classifier weights for each class, the model retains strong discriminative power and enables feature representations aligned with the identity label space. QAConv~\cite{liao2020qa} goes beyond representation learning and explicitly analyzes how person images are matched via deep feature maps. Unlike conventional DG-ReID models that rely on fixed convolution filters, QAConv dynamically constructs query-adaptive convolution kernels from local regions of the query feature map. These kernels are then convolved with the gallery feature map to compute a similarity map that captures fine-grained correspondences. A global max pooling operation is applied to extract the most discriminative matches, enabling precise query-to-gallery alignment. In addition, feature maps corresponding to each class are stored in a class-specific memory bank, where the most recent feature map of each class is directly replaced. This direct replacement strategy preserves the latest representation of each identity, which is more suitable for maintaining identity-specific details. Similar to DIMN~\cite{song2019dimn}, QAConv also supports memory updates via exponential moving average (EMA). However, since EMA performs a running average over multiple batches, it may blur local details in the feature map. As a result, class-specific local patterns could be smoothed out in memory, which may degrade matching precision in local correspondence tasks.To facilitate local part-level matching, PAT~\cite{ni2023pat} maintains a momentum-updated memory bank that stores three part-wise features per image across the training set. This memory supports cross-ID similarity learning and proxy tasks by enabling comparisons between current features and accumulated representations.

\begin{equation}
w_{p_i}^j \leftarrow 
\begin{cases}
f\left(x_{p_i}^j\right), & \text{if } t = 0 \\
(1 - m) \cdot w_{p_i}^j + m \cdot f\left(x_{p_i}^j\right), & \text{if } t > 0,
\end{cases}
\label{eq:memory_momentum_update}
\end{equation}
as shown in Eq.~\ref{eq:memory_momentum_update}, each memory entry $w_{p_i}^j$ is initialized with the first extracted part feature and updated over training epochs using exponential moving average. Here, $f(\cdot)$ is the feature extractor, $x_{p_i}^j$ is the $i$-th part of sample $j$, $t$ is the current epoch, and $m$ is the momentum coefficient. This strategy ensures stable aggregation of reusable local visual patterns across identities. Subsequently, Part-guided Self-Distillation (PSD) compares each part-level feature with all stored features in the memory bank and selects the top-$k$ most similar patches as positive pairs. The corresponding identity labels are used to construct soft labels, which are then employed to supervise the learning of global representations. Unlike conventional self-distillation methods that rely on the model's own predictions, PSD treats the soft labels derived from a proxy task as the teacher signal. This encourages learning based on local appearance similarities and reduces reliance on domain-specific information.

All three methods share the common strategy of progressively refining and aligning feature representations during training through a memory bank that stores class-specific samples across batches. At inference time, only the refined representations are used for identity matching without further reliance on the memory. However, since these methods store additional data during training, they inherently require memory proportional to the number of classes $C$, feature dimension $D$, and spatial size $H \times W$. This becomes particularly problematic in DG-ReID, where multi-source training is performed using three or more datasets, leading to potential GPU memory bottlenecks. Moreover, the memory bank update speed may degrade as the number of classes increases, and class imbalance can cause under-updated entries. Finally, as the memory bank is not involved during inference, its contribution to domain generalization is indirect and cannot be explicitly verified.

\subsection{Meta-learning-based}
\label{subsec3-4}
Training strategies are critical in Domain Generalizable Person Re-identification (DG-ReID), as relying solely on conventional representation learning approaches poses fundamental limitations. Specifically, when models are trained only on labeled source domains, they tend to overfit to domain-specific features such as background, brightness, or camera styles, which may not generalize well to unseen domains. To overcome this challenge, recent DG-ReID methods incorporate advanced training strategies that simulate target domain conditions during training. A prominent approach is meta-learning~\cite{finn2017maml}, which structures training into episodic tasks that mimic domain shifts. Model-Agnostic Meta-Learning (MAML)~\cite{finn2017maml} seeks optimal initial parameters that enable fast adaptation to new tasks, and has proven effective in few-shot learning by simulating task shifts. In the context of domain generalization (DG), MLDG~\cite{li2018metalearning} extends MAML by simulating domain shifts through a meta-train/meta-test split among the source domains. However, MLDG is primarily tailored for classification tasks under closed-set label assumptions, which limits its direct applicability to person re-identification. DG-ReID presents a unique challenge as it involves open-set label distributions and cross-camera variations, requiring not only task-level adaptation but also robust feature-level generalization across domains. Instead of treating each domain as a separate task, DG-ReID methods formulate meta-learning at the domain level, emphasizing inter-domain alignment, dynamic adaptation, and feature robustness.

\begin{algorithm}[t]
\small
\caption{Meta-Learning Training Framework}
\label{alg:meta}
\begin{algorithmic}[1]
\REQUIRE Source domains $\mathcal{D} = \{\mathcal{D}_1, \dots, \mathcal{D}_K\}$; parameters $\theta_f$ (feature extractor), $\phi$ (classifier), $\rho$ (auxiliary module for meta-test)\\ $\mathcal{D}_{mtr}$: meta-train domains, $\mathcal{D}_{mte}$: meta-test domains \\ $X_B$: batch for base update, $X_S$: meta-train batch, $X_T$: meta-test batch
\ENSURE Trained Feature extractor $f_{\theta_f}(\cdot)$ and classifier $g_\phi(\cdot)$
\STATE Initialize $\theta_f$, $\phi$, $\rho$
\FOR{each iteration}
    \STATE \textbf{Base Update:}
    \STATE \hspace{1em} Sample $X_B$ from $\mathcal{D}$
    \STATE \hspace{1em} Compute $\mathcal{L}_{base} = \mathcal{L}_{\text{cross-entropy}}(X_B) + \mathcal{L}_{\text{triplet}}(X_B)$           // Eq.~\ref{eq:triplet}
    \STATE \hspace{1em} Update $\theta_f$, $\phi$ using $\nabla \mathcal{L}_{base}$
    \STATE \textbf{Domain-Level Sampling:}
    \STATE \hspace{1em} Partition $\mathcal{D}$ into $\mathcal{D}_{mtr}$ and $\mathcal{D}_{mte}$ s.t. $\mathcal{D}_{mtr} \cap \mathcal{D}_{mte} = \emptyset$
    \STATE \textbf{Meta-Train:}
    \STATE \hspace{1em} Sample $X_S$ from $\mathcal{D}_{mtr}$
    \STATE \hspace{1em} Compute $\mathcal{L}_{mtr} = \mathcal{L}_{\text{cross-entropy}}(X_S) + \mathcal{L}_{\text{triplet}}(X_S)$          // Eq.~\ref{eq:triplet}
    \STATE \hspace{1em} Update $\theta_f$ using $\nabla \mathcal{L}_{mtr}$
    \STATE \textbf{Meta-Test:}
    \STATE \hspace{1em} Sample $X_T$ from $\mathcal{D}_{mte}$
    \STATE \hspace{1em} Compute $\mathcal{L}_{mte} = \mathcal{L}_{\text{triplet}}(X_T)$
    \STATE \hspace{1em} Update $\rho$ using $\nabla \mathcal{L}_{mte}$
\ENDFOR
\end{algorithmic}
\end{algorithm}

\paragraph{Domain-level learning}
Given the substantial variability among source domains in DG-ReID, recent approaches have incorporated domain-level meta-learning strategies to better capture and mitigate domain shifts. DMG-Net~\cite{bai2021person30k} introduces a dual-meta learning framework consisting of a meta-generalization training procedure and a meta-discrimination loss. The training data is partitioned based on camera IDs, where each camera is treated as a distinct pseudo-domain. In each meta-training episode, cameras are randomly divided into support and query sets. The model first computes the loss on the support camera set and performs gradient descent to obtain adapted parameters. These parameters are then evaluated on the query camera set to compute a second loss, which is used to update the model. In essence, the method applies MAML~\cite{finn2017maml} meta-learning at the camera level to simulate domain shifts and enhance generalization across domains. Memory-based Meta-Learning for Domain Generalization (M$^3$L)~\cite{zhao2021memory} maintains a memory bank that tracks domain-specific feature statistics and their relevance to unseen domains. This allows the model to learn how different domains relate and generalize better to unknown environments by referencing these inter-domain relations during training. In contrast, Meta-Distribution Alignment (MDA)~\cite{ni2022mda} proposes a framework that aligns the distributions of multiple source domains through meta-optimization. Rather than relying on meta-task structures, MDA treats the source domains as an ensemble and directly enforces consistency in their feature distributions to ensure the learned representations remain domain-agnostic. This process is formalized in \hyperref[alg:meta]{Algorithm 1}. Here, $\theta_{f}$ denotes the parameters of the feature extractor, $\phi$ refers to the parameters of the classifier head, and $\rho$ represents the parameters of the auxiliary module used specifically in the meta-test phase. In each training iteration, the base loss is computed, which consists of the sum of the standard cross-entropy loss and the triplet loss. The triplet loss is formulated as:
\begin{equation}
\label{eq:triplet}
\mathcal{L}_{\text{triplet}} = \max\left(0, \|f_a - f_p\| - \|f_a - f_n\| + \text{margin} \right),
\end{equation}
where $f_a$, $f_p$, and $f_n$ denote the features of the anchor, positive, and negative samples, respectively. The margin enforces a minimum distance between the anchor-positive and anchor-negative pairs. This base loss serves as the fundamental objective in ReID to learn discriminative identity features. Next, domain-level sampling is performed by splitting the training domains into a meta-train set $D_{\text{mtr}}$ and a meta-test set $D_{\text{mte}}$, ensuring no overlap between them. These two sets together cover all source domains used in training. During the meta-train phase, the model performs the same operations as the base update but only on $D_{\text{mtr}}$, updating $\theta_{f}$ accordingly. In the subsequent meta-test phase, only the triplet loss is used to update $\rho$. This design simulates the test environment on an unseen domain, where identity labels are unavailable. Therefore, cross-entropy loss is not applied in this stage, as $\phi$ is already optimized for the meta-train domains and may lead to domain-specific bias if reused. Both M$^3$L and MDA move beyond conventional meta-learning paradigms by leveraging domain-level supervision and optimization to reduce domain gaps. Additionally, several recent approaches~\cite{choi2021metabin, xu2022mimic, zhang2022adaptive, dai2021ramoe} simulate domain generalization by constructing meta-test sets during training, forcing the model to generalize across synthetic domain shifts. 

MetaBIN~\cite{choi2021metabin} and ACL~\cite{zhang2022adaptive} aim to reduce domain gaps through feature variance suppression. However, these normalization-based methods lack explicit guidance on which domain statistics should be suppressed or preserved. To address this limitation, meta-learning is introduced to provide supervision based on performance on meta-test domains. This enables the model to identify domain-specific statistics that hinder generalization and to selectively suppress them. As a result, meta-learning alleviates the risk of removing ID-discriminative information--a common drawback in normalization-based approaches--and allows the normalization ratio to be explicitly controlled beyond standard classification objectives. Similarly, META~\cite{xu2022mimic} and RaMoE~\cite{dai2021ramoe} adopt mixture-of-experts architectures to promote representation diversity through expert disagreement. While this design captures a wide range of domain variations, it provides no explicit criterion for selecting which expert to use under which domain condition. Consequently, the model may overfit to specific experts or suffer from domain collapse due to suboptimal expert selection. These hybrid frameworks benefit from their domain-aware and input-adaptive components--such as normalization statistics or expert selection mechanisms--that can be guided by meta-learning objectives. In other words, they contain tunable components that respond to domain shifts, making them highly compatible with meta-learning, which provides optimization signals based on unseen domains. Nonetheless, the effectiveness of meta-learning critically depends on the quality of the meta-train/meta-test split. If the two domains are too similar, the meta-test loss provides little useful supervision, leading to trivial solutions and potential overfitting. Conversely, if the domains are too dissimilar, the model may struggle to identify transferable patterns, resulting in unstable feature representations or over-specialization to a single domain--ultimately leading to sub-optimal generalization performance.

\subsection{Data-driven learning-based}
\label{subsec3-5}
Data-driven learning-based methods aim to enhance domain generalization by diversifying training data. In DG-ReID, collecting person images from real-world surveillance footage raises substantial privacy concerns and annotation costs. Moreover, the generalization capability of DG-ReID models heavily depends on the availability of diverse domains, making the reliance on manually labeled datasets a limiting factor. To address these challenges, researchers have explored three major directions: synthetic data generation, unlabeled data learning, and data augmentation. Each approach offers a complementary perspective for increasing domain variability while mitigating the reliance on exhaustive manual annotations.

\paragraph{Synthetic Data Generation}
Synthetic datasets offer a privacy-preserving and cost-efficient alternative to manually annotated data, particularly in surveillance and person re-identification tasks where identity privacy is a critical concern. Recent reviews~\cite{nikolenko2021synthetic, delussu2024synthetic} highlight that synthetic data not only reduces annotation cost but also enables controlled variation of pose, lighting, occlusion, and camera viewpoint. These properties are especially beneficial in surveillance applications, where domain shifts often arise from environmental variability and privacy regulations. Delussu et al.~\cite{delussu2024synthetic} further provide a taxonomy of synthetic generation techniques--including GAN-based methods, computer graphics engines, and image composition pipelines--and analyze their effectiveness across specific surveillance tasks such as crowd counting, tracking, and person re-identification. Recent methods have explored the use of synthetic data specifically for person re-identification, leveraging its controllable properties to simulate diverse appearance variations and domain conditions. GCL+~\cite{chen2022gcl} utilizes a 3D mesh-based generator to disentangle identity-related and -unrelated features, enabling the synthesis of person images with diverse poses and viewpoints. In a different approach, applies a GAN to transfer the style of the source domain onto target images~\cite{putzu2023specialise}, thereby reducing domain shifts. By combining features from both original and style-transferred images, the method achieves more consistent ranking performance across diverse domains. Earlier synthetic datasets such as SyRI~\cite{bak2018domain} and PersonX~\cite{sun2019dissecting} simulated virtual humans using controlled rendering environments to facilitate domain variation. While these datasets enabled initial explorations of domain shift in person re-identification, their diversity was limited due to constrained character models, simplistic environments, and limited variations in pose or clothing. To address these shortcomings, ClonedPerson~\cite{wang2022cloning} introduces a more sophisticated pipeline that clones identity-specific appearance from real ReID images into synthetic 3D environments. Specifically, it extracts clothing textures from real images using a keypoint-based part alignment mechanism and maps them onto rigged 3D avatars through UV texture mapping~\cite{chang2006modeling}. These avatars are then rendered under various poses, camera angles, and illumination conditions within photo-realistic synthetic environments. Unlike prior datasets that simulate synthetic identities from scratch, ClonedPerson ensures that each synthetic subject directly corresponds to a real-world identity. This real-to-synthetic identity preservation enables more realistic training scenarios while retaining the controllability of synthetic data, making it a valuable asset for domain generalization research. Despite such advances, synthetic datasets still face challenges in bridging the gap with real-world distributions. Texture realism, background complexity, and motion dynamics often remain insufficient, limiting generalization when models are trained solely on synthetic data. To mitigate this, recent studies have explored hybrid training paradigms—such as pretraining on synthetic data followed by fine-tuning with limited real samples—or integrated domain adaptation techniques that reduce synthetic-to-real discrepancies during training. As high-quality synthetic datasets continue to evolve, they offer a promising pathway for scalable, privacy-compliant data-driven domain generalization in ReID.

\begin{figure}[t!]
  \centering
  \begin{tikzpicture}
    \node[inner sep=0pt] (img) at (0,0) {\includegraphics[width=\linewidth]{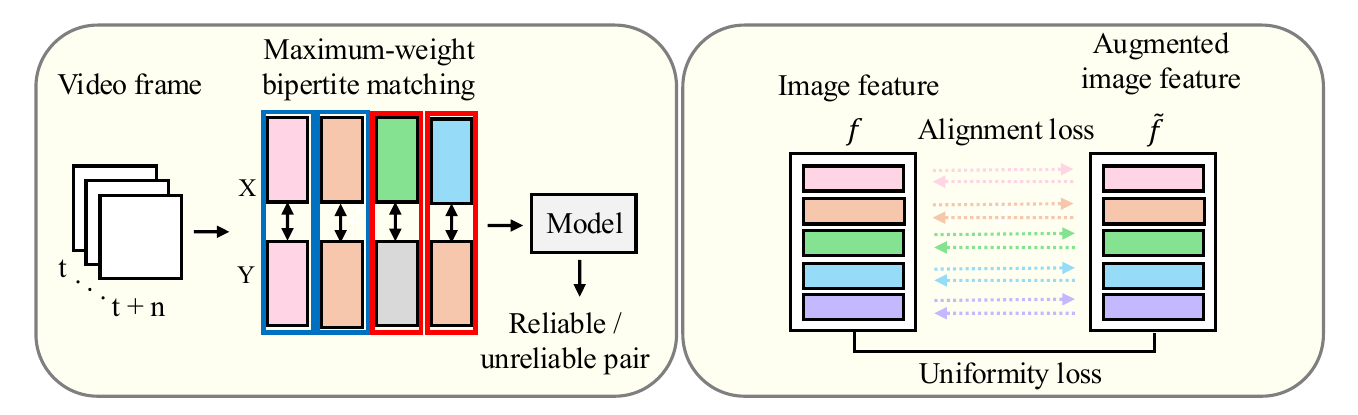}};
    \node[anchor=north west] at ([xshift=90pt,yshift=-115pt]img.north west) {\footnotesize(a)};
    \node[anchor=north east] at ([xshift=-90pt,yshift=-115pt]img.north east) {\footnotesize(b)};
  \end{tikzpicture}
  \vspace{-25pt}
  \caption{Overview of data-driven methods for DG-ReID. (a) Unlabeled data learning using frame-wise bipartite matching~\cite{dou2023identity}, where blue boxes indicate correct matchings and red boxes indicate incorrect ones. (b) Augmentation-based learning with alignment and uniformity losses~\cite{cho2024alignment}.}
  \vspace{-10pt}
  \label{fig:data_methods}
\end{figure}

\paragraph{Unlabeled data learning}
To bridge the gap between synthetic and real data, recent works leverage unlabeled data from large-scale, diverse, real-world video datasets. ISR~\cite{dou2023identity}, for instance, utilizes YouTube-VOS~\cite{xu2018youtubevos}, which features natural motion and appearance changes across time. Unlike conventional ReID datasets that rely on static surveillance cameras, YouTube-VOS provides dynamic and varied visual conditions, helping models learn representations robust to temporal variation. As shown in Fig.~\hyperref[fig:data_methods]{5} (a), ISR~\cite{dou2023identity} extracts two frame sets $X$ and $Y$ at time $t$ and $t+n$ from a video sequence. Without using labels, identity associations are discovered through bipartite matching, even across temporally distant frames. To ensure robust learning, ISR introduces a reliability-guided contrastive loss that suppresses noisy pseudo-positive samples, enabling the model to learn temporally coherent identity representations and improving training stability. This approach highlights the potential of unlabeled video data to reduce domain gaps while preserving semantic consistency. However, learning from unlabeled data also presents challenges. Pseudo-labeling based on visual similarity can be error-prone, especially across large temporal gaps or under occlusion, leading to the propagation of noisy identity associations. Although techniques such as reliability-guided contrastive loss aim to mitigate this issue, the learning process remains sensitive to false positives. Furthermore, such methods generally rely on pre-trained models or auxiliary modules for feature initialization or filtering, which may introduce domain biases unless carefully tuned. Overall, leveraging unlabeled video data presents a promising direction for bridging the synthetic-to-real gap, but requires robust association strategies and uncertainty-aware learning to fully realize its potential in domain generalizable person re-identification.

\paragraph{Augmentation}
Data augmentation remains a widely used and effective method for improving generalization, especially when labeled data is limited. Beyond traditional techniques like random flipping and cropping, recent DG methods explore adversarial~\cite{volpi2018adversarial, zhao2020maximumentropy} and style-based~\cite{zhong2022adversarial} augmentations to simulate domain shifts. However, in ReID tasks, preserving person-specific features is critical; overly aggressive augmentations such as random erasing~\cite{zhong2020randomerasing} can inadvertently remove identity-relevant cues like clothing patterns or accessories, thereby reducing the model's ability to distinguish individuals across domains. To address this trade-off, BAU~\cite{cho2024alignment} introduces a novel learning strategy that balances alignment and uniformity in the feature space. As shown in Fig.~\hyperref[fig:data_methods]{5} (b), alignment ensures that samples of the same identity are closely clustered, promoting discriminative identity representation. Conversely, uniformity encourages feature vectors to be evenly distributed across the representation space, enhancing generalization by preventing feature collapse. The method jointly optimizes both objectives, enabling models to learn features that are both distinctive and domain-invariant. Overall, augmentation-based methods are powerful tools for domain generalization due to their simplicity and flexibility. However, they require careful calibration to preserve fine-grained identity cues and to prevent the learning process from being misled by semantically inconsistent or overly distorted samples. Methods like BAU exemplify how augmentation can be combined with feature-level constraints to mitigate these risks and maximize generalization potential.

From the perspective of training strategies, meta-learning and data-driven methods adopt fundamentally different philosophies for addressing domain shift. Meta-learning focuses on how to learn by simulating domain shifts through episodic training and optimizing the model to generalize across unseen domains. In contrast, data-driven methods--such as synthetic data generation, unlabeled data learning, and augmentation--focus on what to learn by manipulating the input data itself to mitigate domain shifts or to expose the model to more diverse conditions. Meta-learning provides an optimization framework that is explicitly aligned with the goal of domain generalization, often leading to more principled feature learning. However, it typically requires more complex training protocols, such as domain-aware sampling and multi-level updates, which can be computationally intensive and sensitive to the quality of domain splits. On the other hand, data-driven methods offer a more straightforward path to generalization by enriching or adjusting the training data. While they are often easier to implement and scale, they also carry inherent limitations, such as synthetic-to-real domain gaps, label noise, or style overfitting, which can negatively affect representation quality if not carefully managed. These two paradigms offer complementary advantages: meta-learning targets robust learning mechanisms, whereas data-driven methods aim to broaden the data landscape. A hybrid design that unifies their strengths may offer a more comprehensive solution to domain shift in person re-identification.

\subsection{CLIP-based}
\label{subsec3-6}
Most existing DG-ReID methods such as normalization-based, mixture-of-experts-based, memory-based, and meta-learning-based approaches focus solely on learning identity-discriminative features from visual representations. With the emergence of pre-trained vision-language models like CLIP~\cite{radford2021clip}, there has been growing interest in leveraging text features alongside image features for identity matching. CLIP has recently demonstrated superior performance across various downstream tasks, including image classification~\cite{ma2023prod} and semantic segmentation~\cite{zhou2023zegclip}, motivating its application to ReID as well. CLIP-ReID~\cite{li2023clipreid} investigates whether fine-tuning the image encoder of CLIP alone can yield competitive performance across diverse ReID benchmarks. CLIP-ReID adopts a two-stage strategy: it first learns task-specific text prompts via contrastive learning with frozen encoders, then fine-tunes the image encoder using standard ReID losses while keeping the learned prompts fixed. While CLIP-ReID successfully adapts the CLIP architecture for fine-grained ReID, it remains suboptimal in domain generalization settings. The learned text prompts are optimized based solely on the training domain and become entangled with its specific styles and biases. As a result, they fail to provide effective semantic guidance when encountering unseen domains. This limitation motivates the emergence of CLIP-based DG-ReID methods to enhance cross-domain generalization through more adaptive or regularized training strategies.

\paragraph{Text-Guided Learning}

To address the limitations of CLIP-ReID~\cite{li2023clipreid} in domain generalization, CLIP-FGDI~\cite{zhao2025clipfgdi} proposes a bidirectional guidance strategy using two types of prompts: domain-invariant prompts and domain-relevant prompts. The domain-invariant prompt adopts the same template as CLIP-ReID, while the domain-relevant prompt includes dataset information in the form of ``A photo of a $[X]_1 [X]_2 \ldots [X]_M$ person from $[D]_1[D]_2..[D]_N$ dataset.'' The model is trained in three stages. The first stage, only the image encoder is updated for a few epochs (\textit{e.g.}, 3 epochs) to warm up the image representations for ReID. And the second stage, both the image and text encoders are frozen, and the prompts are optimized. The domain-invariant prompts are trained with a domain classifier and a gradient reversal layer (GRL) to make the resulting text features domain-agnostic. The last stage, the learned prompts are fixed, and the image encoder is fine-tuned using ID classification and triplet loss, following the same strategy as CLIP-ReID. The key difference is that CLIP-FGDI introduces an additional prompt-level triplet loss:
\begin{equation}
\mathcal{L}_{\text{apn}} = \max\left(0, \text{Sim}(f^I_a, f^T_p) - \text{Sim}(f^I_a, f^T_n) + m\right),
\label{eq:apn}
\end{equation}
where $f^I_a$ denotes the anchor image feature, $f^T_p$ is the text feature of the domain-invariant prompt (positive), $f^T_n$ is that of the domain-relevant prompt (negative), and $m$ is the margin (set to 0.3). This loss encourages the image feature to align more closely with the domain-invariant prompt while pushing it away from the domain-specific one. CLIP-FGDI achieves better generalization performance than CLIP-ReID on unseen target domains, demonstrating the effectiveness of using prompt-based textual guidance for DG-ReID. However, the method has a notable limitation: domain tokens are based on dataset names (\textit{e.g.}, Market-1501~\cite{zheng2015scalable}), and empirical results show that using a single domain token ($N=1$) yields the best performance. This suggests that the model may implicitly rely on dataset-specific cues such as background or lighting. In real-world deployment scenarios, where domain names may be ambiguous or unavailable, generating appropriate domain tokens can become non-trivial, limiting the scalability and general applicability of the method. CLIP-DFGS~\cite{zhao2024clipdfgs} proposes a Depth-First Graph Sampler (DFGS) that leverages text tokens output from the text encoder as semantic anchors to align image samples across the entire batch in accordance with their corresponding identities. Traditional sampling strategies such as the PK sampler~\cite{hermans2017defense} randomly select \( P \) classes and sample \( K \) instances per class in each mini-batch. While this ensures sample diversity, it fails to consistently provide hard samples that are beneficial for model learning. In contrast, the graph sampler~\cite{liao2020qa}, as introduced in Section~2.3 in the context of QAConv, is tightly coupled with specific model architectures and thus not readily applicable to general ReID frameworks. To overcome these limitations, DFGS first constructs a feature similarity graph by computing a pair-wise distance matrix from the features extracted by the text encoder. This distance matrix is calculated based on Euclidean distances between text embeddings, from which top-\( k \) nearest neighbors are selected to define graph edges. Then, a depth-first search (DFS) is performed over the graph to sequentially traverse classes that are semantically similar yet distinct--\textit{i.e.}, hard negatives--forming a mini-batch of hard samples. Following this, image samples are selected based on their alignment with the corresponding text anchors. Similar to CLIP-FGDI, CLIP-DFGS employs the two-stage structure of CLIP-ReID, where prompt tokens are first learned and then used to align image features with the text domain. By leveraging CLIP’s cross-modal alignment capability, DFGS enables graph construction and sampling in both the image and text embedding spaces. This allows for semantic-level hard sample mining guided by text anchors, improving generalization to unseen domains. However, no specific strategy is proposed to prevent the text prompts from overfitting to the training domains, which is critical for domain generalizable ReID. In addition, the method requires computing pair-wise distances between all text features, resulting in a computational cost of \( \mathcal{O}(N^2) \) when the number of text prompts becomes large.

\subsection{Others}
\label{subsec3-7}
In this section, we introduce several DG-ReID approaches that are not encompassed by the primary methodological categories previously discussed--namely, normalization-based, mixture-of-experts-based, memory-based, meta-learning-based, and CLIP-based methods--but nonetheless present promising directions for enhancing domain generalization. These approaches address DG challenges through alternative methodological frameworks, including the resolution of task interference via gradient alignment and the promotion of domain diversity under privacy constraints in federated learning scenarios. Although these methods remain relatively underexplored, they provide complementary insights and hold significant potential to advance future DG-ReID research.

\paragraph{Gradient-Alignment}
PAOA~\cite{li2024paoa} introduces a gradient alignment framework tailored for DG-ReID. The method couples the primary task of instance classification with an auxiliary saliency detection task. During the forward pass, backbone features are decomposed into two branches: one dedicated to identity classification and the other to saliency detection. The saliency branch is supervised using weak saliency labels generated by a pre-trained detector~\cite{zhao2019pfan}, with an $L_1$ loss. Crucially, during backpropagation, PAOA enforces orthogonality between the gradients of the auxiliary and primary tasks. This ensures that the saliency-based auxiliary supervision does not interfere with identity-relevant gradient signals. By aligning gradients in this manner, PAOA facilitates the disentanglement of domain-specific noise (\textit{e.g.}, background, scale, or camera view) from identity features, improving the generalization capacity of the learned representation. To further enhance adaptability, PAOA+~\cite{li2024paoa} extends the framework to support test-time optimization. While all components are trained jointly, PAOA+ enables the auxiliary branch to continue adapting at deployment using a small number of unlabeled target-domain samples. Saliency maps are regenerated via the pre-trained detector, and only the auxiliary branch is updated to refine feature decomposition. This lightweight test-time adaptation enhances robustness to domain shifts without relying on identity labels or domain annotations during training. While PAOA effectively mitigates gradient conflict between tasks, certain challenges remain. Since the auxiliary supervision relies on saliency detection, it may not always correspond to identity-relevant cues, especially under pose or occlusion variations. Moreover, the approach depends on the quality of a pre-trained saliency detector, which might not generalize well across ReID domains. Lastly, although gradient orthogonality reduces interference, it may not fully disentangle overlapping semantic features, potentially limiting its flexibility in more complex scenarios.

\paragraph{Federated Stylization}
DACS~\cite{dai2023dacs} tackles domain generalization in person re-identification from a federated learning perspective. This setup reflects real-world privacy constraints. Unlike conventional stylization methods that require multi-domain access, DACS introduces a Diversity-Authenticity Co-constrained Stylization framework that locally generates style-diverse yet identity-preserving samples. For each mini-batch of local images, DACS computes channel-wise data statistics and replaces them with learnable statistics $\hat{\mu}$ and $\hat{\sigma}$ from a Style Transformation Model (STM), enabling lightweight stylization without a separate generator. To encourage greater diversity, a distributional distance (2-Wasserstein distance~\cite{he2018wasserstein}) is maximized between the original and stylized feature distributions. To ensure that stylized images remain realistic and semantically valid, DACS leverages two ReID models per client: a local model $f_L$, optimized on local data and thus biased toward domain-specific features, and a local-side global model $f_G$, initialized from the averaged global model. The stylized image $x'$ is encouraged to be harder for the global model but easier for the local model, promoting domain shift while preserving identity information. The authenticity loss is defined as:
\begin{multline}
\mathcal{L}_{\text{au}}(x; \phi) =
\text{Softplus}(H(f_{global}(x)) - H(f_{global}(x'))) + \\
\text{Softplus}(H(f_{global}(x')) - H(f_{local}(x')))
\end{multline}
Here, $x$ and $x'$ denote the original and stylized images, respectively. $f_{global}(\cdot)$ and $f_{local}(\cdot)$ represent the ``local-side global model'' and ``local model.'' $H(\cdot)$ denotes the entropy of the model’s prediction, used to approximate prediction uncertainty. $\text{Softplus}(\cdot) = \ln(1 + e^x)$ is a smooth activation function for stable optimization. While DACS offers an efficient and privacy-compatible solution, it may face certain limitations. Since stylization is performed within each local domain, the diversity of generated samples is inherently constrained. Additionally, relying solely on first-order and second-order statistics may limit the ability to simulate complex domain characteristics, such as background clutter or camera-specific effects, potentially reducing the generalization benefit.

\section{Cross-Task Perspective: DG-ReID and VI-ReID}
\label{sec4}
This section examines how modules developed for DG-ReID are also applied in related tasks that address similar domain gap challenges. By comparing these approaches, we highlight the applicability and potential extension of DG-ReID techniques, and provide a clearer understanding of how they handle domain-specific variations. Specifically, we investigate Visible-Infrared Re-ID (VI-ReID), which aims to match identities across visible and infrared modalities, where both modalities are available per identity. In contrast, DG-ReID assumes each identity appears in only one domain and focuses on generalizing to unseen domains. This structural difference leads to distinct modeling approaches. Despite this, both tasks share a core challenge--bridging heterogeneous distributions across domains or modalities--commonly referred to as the domain gap. To address this, both DG-ReID and VI-ReID adopt overlapping learning strategies, including adversarial learning, memory-based modeling, bipartite graph matching, normalization techniques, and knowledge distillation. In this section, we explore their conceptual and methodological intersections. Section~\ref{subsec4-1} focuses on understanding the modality gap in VI-ReID, highlighting how it differs from the domain gap addressed in DG-ReID. Section~\ref{subsec4-2} introduces representative VI-ReID methods that share underlying principles with DG-ReID approaches. This analysis reveals transferable insights and encourages cross-task research.

\subsection{Understanding the Modality Gap in VI-ReID}
\label{subsec4-1}
Visible-Infrared Re-Identification (VI-ReID) aims to identify the same individual across visible and infrared images, which are typically captured under different lighting conditions, visible during the day and infrared at night. This task is practically important in 24-hour surveillance scenarios, where poor illumination at night makes visual recognition particularly challenging. The core difficulty lies in the modality gap, referring to the substantial appearance discrepancies between heterogeneous pedestrian images from different sensing modalities. To address this gap, existing works primarily follow two directions: image-level and feature-level solutions. Image-level approaches attempt to unify modality appearance via image translation or style generation techniques, enabling models to operate in a single modality space. On the other hand, feature-level methods aim to learn a modality-shared embedding space that minimizes cross-modality discrepancies while preserving identity semantics. While VI-ReID and DG-ReID differ in the nature of the domain gap--modality discrepancy versus environment-induced variation--they share the common objective of learning robust representations that generalize across diverse domains.

\subsection{Domain Generalization Techniques Applied to VI-ReID}
\label{subsec4-2}
Visible-Infrared Re-Identification (VI-ReID) often incorporates various generalization modules to mitigate the modality gap and extract robust identity-discriminative features. In this subsection, we examine representative VI-ReID methods that adopt generalization techniques either directly derived from or conceptually aligned with those used in DG-ReID. We analyze the shared principles and methodological distinctions, highlighting how DG-oriented strategies have been adapted to the cross-modality setting of VI-ReID.

CMT~\cite{jiang2022cross} proposes the Cross-Modality Transformer, which integrates both modality-level and instance-level alignment modules for VI-ReID. The instance-level module performs query-adaptive feature modulation by generating channel-wise scaling and shifting parameters from the global feature of query via an MLP. These parameters are then used to apply affine transformations to gallery features, enhancing relevant channels while suppressing irrelevant ones. This mechanism similar to normalization-based methods in Section~\hyperref[subsec3-1]{3.1}, such as IN~\cite{jin2020style} and learnable combinations of IN and BN~\cite{zhang2022adaptive, choi2021metabin}, which also operate at the channel level. From the perspective of DG-ReID, CMT shares the goal of learning domain-invariant representations. However, it differs in mechanism and supervision target, as it dynamically adapts gallery features based on each query at test time, rather than relying on global domain-level statistics. This design offers fine-grained alignment, but may raise concerns regarding model consistency under query variation. Furthermore, unlike many DG approaches, CMT does not explicitly disentangle domain-invariant and domain-specific components. Overall, while conceptually aligned with DG principles, CMT reflects a task-specific adaptation tailored to the cross-modal setting of VI-ReID.

OTLA-ReID~\cite{wang2022optimal} introduces a Discrepancy Elimination Network (DEN), which consists of a modality classifier and a Gradient Reversal Layer (GRL). The modality classifier predicts the modality source of a given feature, while the GRL inverts gradients during backpropagation, encouraging the feature extractor to learn modality-invariant representations. This mechanism is conceptually similar to the domain confusion strategy used in CLIP-FGDI~\cite{zhao2025clipfgdi}, as discussed in Section~\hyperref[subsec3-6]{3.6}, where a domain classifier and GRL are employed to suppress domain-specific cues. While both approaches employ the same GRL-based architecture, they differ in how the domain is defined. DG-ReID typically considers different datasets collected under varied environmental conditions as separate domains, whereas VI-ReID defines the visible and infrared modalities within a single dataset as distinct domains. Nevertheless, GRL proves effective in both cases, as its underlying objective--discouraging the encoder from learning domain- or modality-specific information--remains applicable across different forms of distributional discrepancy. However, the implications of this discrepancy differ. In VI-ReID, domain confusion is binary and well-defined, while DG-ReID involves multiple domains with more complex and entangled biases. Consequently, GRL-based classifiers in DG-ReID may require more sophisticated designs to handle greater variation. Furthermore, whereas VI-ReID pursues cross-modal invariance within a single dataset, DG-ReID targets broader generalization across unseen domains, affecting both the robustness and scalability of GRL-based approaches.

GUR~\cite{yang2023towards} introduces Cross-memory Association Embedding (CAE), where modality-specific memory banks for visible and infrared images are maintained. Instance features are compared with entries in both memories to generate association embeddings that preserve modality-specific distributions. Similarly, ACCL~\cite{wu2023unsupervised} employs dual contrastive learning with separate memory banks for each modality, storing cluster centroids that are updated through early-stage clustering. Contrastive learning then pulls features toward their assigned centroids and pushes them away from others within the same modality. These mechanisms are conceptually related to memory-based approaches discussed in Section~\hyperref[subsec3-3]{3.3}, which utilize memory banks to improve representation learning. Despite the structural resemblance, memory usage serves different purposes in the two tasks. VI-ReID benefits from modality-specific memories that preserve the distinct distributions of visible and infrared data, which is particularly effective due to the clear and fixed nature of modality boundaries. However, this design may not transfer well to DG-ReID, where domain boundaries are often ambiguous and confounded with factors like background or camera view. In such cases, modality-specific memories may reinforce, rather than mitigate, domain bias, highlighting the importance of clearly defined domains for effective memory-based strategies.

MMM~\cite{shi2024multi} introduces a Multi-Memory Learning strategy that captures fine-grained modality variations by maintaining multiple sub-memories per identity, each representing distinct intra-ID characteristics. Cross-modal sub-memories are aligned using weighted bipartite graph matching. Similarly, CAL~\cite{wu2023learning} performs bipartite matching between modality-specific query sets based on attention-weighted similarity. While these methods are structurally similar to the bipartite matching used in ISR~\cite{dou2023identity} (Section~\hyperref[subsec3-5]{3.5}), their objectives differ. VI-ReID uses matching to align fine-grained modality-specific patterns under identity supervision, whereas DG-ReID applies it to unlabeled domains to learn modality-agnostic features. The fixed identity supervision in VI-ReID enables more direct part-level alignment across modalities. In contrast, DG-ReID often lacks explicit supervision across domains, making graph-based alignment less precise or harder to scale. This highlights how similar architectures can serve distinct roles depending on the availability of supervision and the granularity of domain discrepancy.

MUN~\cite{yu2023modality} introduces an auxiliary generator consisting of an Intra-Modality Learner (IML) and a Cross-Modality Learner (CML). IML extracts modality-specific features, while CML learns shared representations, both of which guide the main network through knowledge distillation. IDKL~\cite{ren2024implicit} extends this idea by aligning three types of feature affinities--modality-specific, shared, and cross-modal--using KL divergence at the feature level, alongside logit-level distillation between branches. These methods share conceptual ground with the knowledge distillation strategy in PAT~\cite{ni2023pat} (Section~\hyperref[subsec3-3]{3.3}), but differ in architectural intent. While PAT distills global representations for domain-invariant learning, MUN and IDKL employ dual-stream structures to separately model and bridge modality-specific and modality-shared representations. This design is particularly well-suited to VI-ReID, where modality boundaries are fixed and known, allowing explicit decomposition and targeted supervision. In contrast, DG-ReID typically lacks such modality cues, limiting the applicability of fine-grained, branch-specific distillation strategies.

\begin{sidewaystable*}
\caption{Statistics of some commonly used datasets for DG-ReID. \textit{\#ID} denotes the number of unique pedestrians, \textit{\#Image} is the number of images, and \textit{\#Cam} refers to the number of camera views. \textit{Detector} indicates how bounding boxes were extracted, and \textit{Res} specifies whether the image resolution is fixed or varies.}
\vspace{10pt}
\centering
\footnotesize
\setlength{\tabcolsep}{10pt} 
\renewcommand{\arraystretch}{2.0} 
\begin{tabular}{l c c c c c c c c}
\hline
\textbf{Dataset} &  \textbf{Venue} & \textbf{\#ID} &  \textbf{\#Image} & \textbf{\#Cam} & \textbf{Detector} & \textbf{Res} & \textbf{Environment} \\ \hline
CUHK01~\cite{li2012human}   & ACCV'12    & 971   & 3,884  & 2  & hand & fixed & campus\\ 
CUHK02~\cite{li2012human}   & ACCV'12    & 1,816 & 7,264  & 10 & hand & fixed & campus \\ 
CUHK03~\cite{li2014deepreid}  & CVPR'14    & 1,467 & 13,164 & 2 & DPM~\cite{girshick2015deformable} & vary & campus \\ 
Market-1501~\cite{zheng2015scalable} & ICCV'15 & 1,501 & 32,688 & 6 & DPM~\cite{girshick2015deformable} & fixed & campus \\ 
CUHK-SYSU~\cite{xiao2017joint} & CVPR'17   & 8,432 & 18,184 & 1  & hand & fixed & urban scenes \\ 
Duke-MTMC~\cite{zheng2017unlabeled} & ICCV'17 & 1,404 & 36,411 & 8 & DPM~\cite{girshick2015deformable} & fixed & campus \\
MSMT17~\cite{wei2018gan}   & CVPR'18 & 4,101 & 126,441 & 15 & Faster RCNN~\cite{ren2015faster} & vary & indoor, outdoor \\ 
\hline
PRID2011~\cite{hirzer2011classification}  & SCIA'11   & 200   & 1,134  & 2  & hand & fixed & street \\ 
VIPeR~\cite{gray2007evaluating}   & PETS'07   & 632   & 1,264  & 2 & hand & fixed & outdoor \\ 
iLIDS~\cite{li2018unsupervised}   & ECCV'18   & 119   & 476    & 2 & hand & vary & airport \\ 
GRID~\cite{loy2013manifold}   & ICIP'13   & 250   & 1,275  & 8 & hand & vary & subway  \\ 

\end{tabular}
\label{table:dg_reid_datasets}
\end{sidewaystable*}

\begin{figure}[t!]
  \centering
  \begin{tikzpicture}
    \node[inner sep=0pt] (img) at (0,0) {\includegraphics[width=\linewidth]{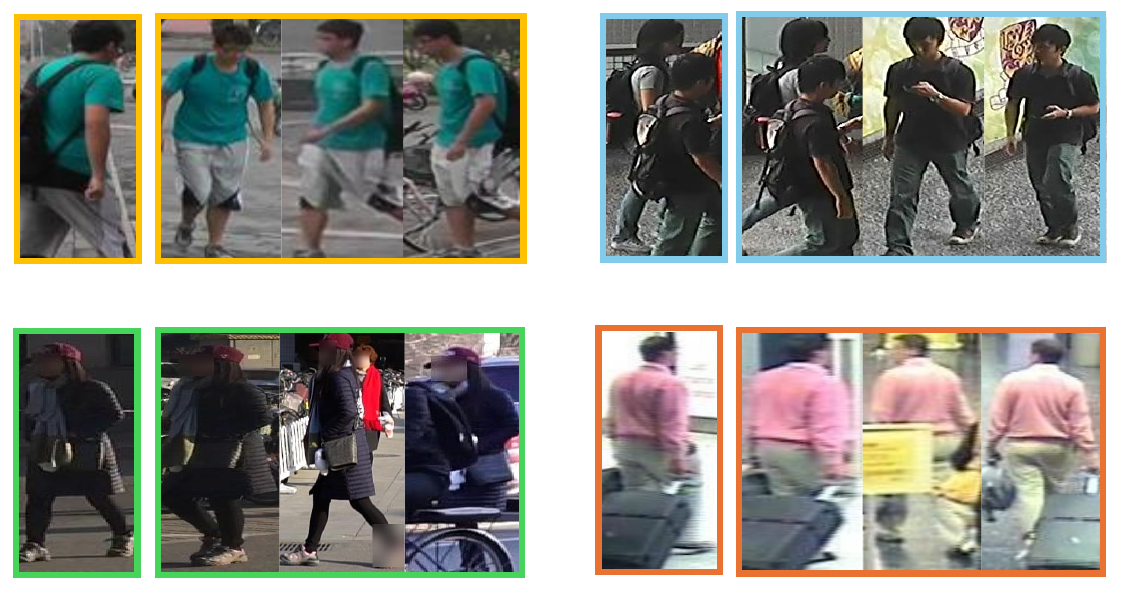}};
    \node[anchor=north west] at ([xshift=80pt,yshift=-90pt]img.north west) {\footnotesize(a)};
    \node[anchor=north east] at ([xshift=-80pt,yshift=-90pt]img.north east) {\footnotesize(b)};
    \node[anchor=north west] at ([xshift=80pt,yshift=-200pt]img.north west) {\footnotesize(c)};
    \node[anchor=north east] at ([xshift=-80pt,yshift=-200pt]img.north east) {\footnotesize(d)};
  \end{tikzpicture}
  \vspace{-25pt}
  \caption{Examples of query (left) and gallery (right) images from different datasets used in DG-ReID evaluation. (Each color corresponds to a different dataset for clarity. Best viewed in color.) (a) Market-1501~\cite{zheng2015scalable}, (b) CUHK03~\cite{li2014deepreid}, (c) MSMT17~\cite{wei2018gan}, and (d) iLIDS~\cite{li2018unsupervised}. While (a) shows relatively clean indoor and campus scenarios, (b) includes label noise. In (c), challenges such as illumination changes and occlusions are observed. (d) iLIDS, captured in an airport environment, exhibits significant domain discrepancy compared to the other datasets. For further dataset statistics and details, refer to Table~\hyperref[table:dg_reid_datasets]{1}.}
  \vspace{-10pt}
  \label{dataset}
\end{figure}

\section{Dataset and Evaluation}
We examine the datasets commonly used in the Domain Generalizable Person Re-identification (DG-ReID) setting and evaluation protocols. 
\label{sec5}
\subsection{Datasets for DG-ReID}
\label{subsec5-1}
DG-ReID benchmarks typically employ 11 widely-used image-based datasets. 
Large-scale datasets such as Market-1501~\cite{zheng2015scalable}, Duke-MTMC~\cite{zheng2017unlabeled}, CUHK01~\cite{li2012human}, CUHK02~\cite{li2012human}, CUHK03~\cite{li2014deepreid}, CUHK-SYSU~\cite{xiao2017joint}, and MSMT17~\cite{wei2018gan} are commonly used for both training and testing. In contrast, small-scale datasets including PRID2011~\cite{hirzer2011classification}, VIPeR~\cite{gray2007evaluating}, iLIDS~\cite{li2018unsupervised}, and GRID~\cite{loy2013manifold} 
are typically used only during the testing phase to evaluate generalization performance. Detailed statistics for these datasets are presented in Table~\hyperref[table:dg_reid_datasets]{1}. \textit{\#ID} indicates the number of unique pedestrian identities, \textit{\#Image} refers to the total number of images, and \textit{\#Cam} denotes the number of camera views. \textit{Detector} specifies the method used to extract pedestrian bounding boxes from the original scene images, and \textit{Res} indicates whether the image resolution is fixed or varies across the dataset. The final column summarizes key aspects of the image environment present in each dataset. In this survey, we highlight test-phase datasets that serve as unseen target domains in DG-ReID benchmarks, as they are central to evaluating cross-domain generalization. Accordingly, we provide detailed descriptions of small-scale benchmarks such as PRID2011, VIPeR, iLIDS, and GRID. For widely-used large-scale datasets like Market-1501, CUHK01/02/03, CUHK-SYSU, and MSMT17, we refer readers to prior comprehensive ReID surveys~\cite{ye2021deep}, as they are already well-documented in the literature.

\vspace{2ex}
\noindent\textit{PRID2011} dataset~\cite{hirzer2011classification} consists of 200 identities captured by two non-overlapping cameras in an outdoor environment. The dataset also includes video sequences for each person. Among the collected identities, only 200 appear in both camera views, enabling cross-view matching. For use in image-based ReID, a fixed number of frames are sampled from each sequence to obtain still images. It offers simple backgrounds and limited diversity, and is typically used only in the testing phase to evaluate cross-domain generalization.

\vspace{2ex}
\noindent\textit{VIPeR} dataset~\cite{gray2007evaluating} is one of the earliest and most challenging ReID datasets, containing 632 identities with one image per camera (two cameras total). The dataset features extreme viewpoint changes and illumination variation. Due to its limited size, it is primarily used for testing generalization performance across domains.

\vspace{2ex}
\noindent\textit{iLIDS} dataset~\cite{li2018unsupervised} is a small-scale person re-identification dataset collected from an airport arrival hall under a multi-camera CCTV network (Fig.~\hyperref[dataset]{6} (d)). It contains 119 identities, each captured from two camera views. The dataset is particularly challenging due to heavy occlusion, varying illumination, and crowded scenes. The images are manually cropped and vary in quality. Because of its difficulty and limited size, iLIDS is typically used only in the testing phase to evaluate the generalization ability of ReID models in domain generalization studies.

\vspace{2ex}
\noindent\textit{GRID} dataset~\cite{loy2013manifold} was collected from 8 disjoint camera views in a subway station, containing 250 probe images and 1,275 gallery images from 250 identities. The images are of relatively low resolution and were manually annotated. Although the dataset was originally captured as video sequences, only still images are provided. GRID presents challenges such as limited camera angles, cluttered backgrounds, and small image sizes. Due to its constrained setting and scale, it is also commonly used as a test-only dataset in DG-ReID scenarios to assess cross-domain robustness.

\subsection{Evaluation Metrics}
\label{subsec5-2}
DG-ReID adopts standard evaluation metrics from the person re-identification literature, namely Rank-1 Accuracy and mean Average Precision (mAP), to measure the retrieval performance of models~\cite{ye2021deep, jin2020style, zhang2022adaptive, xu2022mimic, nie2025normalization, choi2021metabin, dai2021ramoe, zhao2021memory, cho2024alignment}. These metrics are widely accepted due to their intuitive interpretability and effectiveness in capturing both retrieval accuracy and precision. The Cumulative Matching Characteristic (CMC) curve evaluates the probability that the correct gallery image appears within the top-$k$ ranked candidates for a given query. Specifically, Rank-1 Accuracy refers to the probability that the correct match is ranked at the top (\textit{i.e.}, $k=1$), serving as a strict and widely-used indicator of identification performance. Formally, CMC is defined as:

\begin{equation}
\text{CMC}(k) = \frac{1}{N} \sum_{i=1}^{N} I(\text{Rank}_i \leq k), 
\label{dg_reid_dataset}
\end{equation}
Here, $N$ denotes the total number of queries, $\text{Rank}_i$ is the position at which the correct match appears in the sorted gallery list for the $i$-th query, and $I(\cdot)$ is an indicator function that returns 1 if the condition is satisfied and 0 otherwise. CMC evaluates whether a correct match exists within the top-$k$ positions, but it does not consider the full rank distribution of multiple relevant images or the relative precision across them. To address this limitation, mean Average Precision (mAP) is employed as a complementary metric. mAP considers both the rank positions of all relevant matches and the precision at each correct retrieval point. 

\begin{equation}
\text{mAP} = \frac{1}{N} \sum_{i=1}^{N} \text{AP}_i, \quad 
\text{where} \quad 
\text{AP}_i = \frac{1}{m_i} \sum_{k=1}^{n_i} P_i(k) \cdot \text{rel}_i(k),
\label{eq:map}
\end{equation}
In Eq.~\eqref{eq:map}, $\text{AP}_i$ denotes the average precision for query $i$, computed as the mean of precisions at all ranks where a correct match is found. The variable $P_i(k)$ is the precision at rank $k$, and $\text{rel}_i(k) \in \{0, 1\}$ indicates whether the gallery image at rank $k$ is a correct match for query $i$. The final mAP score is obtained by averaging the APs over all queries. This provides a more comprehensive assessment of retrieval quality, especially in cases where multiple positive instances exist per query. Together, CMC and mAP provide a balanced evaluation: CMC captures how early the first correct match appears, while mAP reflects the overall ranking quality and recall. In the context of domain generalizable ReID, where significant variations exist between source and target domains, relying on both metrics is essential for a fair and complete evaluation of generalization performance. These evaluation metrics are also widely adopted across other ReID paradigms beyond DG-ReID, offering a consistent basis for performance comparison in the broader ReID literature~\cite{ding2024msdn, ding2024eraser, ding2024disentangled, ding2025attention}.

\begin{table*}
\caption{Evaluation Protocols. M, C2, C3, and CS denote Market-1501~\cite{zheng2015scalable}, CUHK02~\cite{li2012human}, CUHK03~\cite{li2014deepreid}, and CUHK-SYSU~\cite{xiao2017joint}, respectively.}
\vspace{10pt}
\centering
\footnotesize
\renewcommand{\arraystretch}{2.0} 
\begin{tabular}{l c c}
\hline
\textbf{Protocol} & \textbf{Training Sets} & \textbf{Test Sets} \\ \hline
Protocol-1 & Full-(M+C2+C3+CS) & PRID, GRID, VIPeR, iLIDs \\ \hline
Protocol-2 & M+MS+CS          & C3 \\ 
           & M+CS+C3          & MS \\ 
           & MS+CS+C3         & M \\ \hline
Protocol-3 & Full-(M+MS+CS)   & C3 \\ 
           & Full-(M+CS+C3)   & MS \\ 
           & Full-(MS+CS+C3)  & M \\ \hline
\end{tabular}

\label{table:evaluation_protocols}
\end{table*}

\subsection{Evaluation Protocols}
\label{subsec5-3}
DG-ReID employs three widely-used evaluation protocols--Protocol-1, Protocol-2, and Protocol-3--as summarized in Table~\hyperref[table:evaluation_protocols]{2}. Each protocol is designed to evaluate generalization performance under different levels of domain shift and practical deployment assumptions. In Protocol-1, all available images from the source domains--including both the training and testing subsets--are used for model training. Following the standard evaluation practice in DG-ReID~\cite{jin2020style, xu2022mimic, zhang2022adaptive, choi2021metabin, nie2025normalization, cho2024alignment}, evaluations are conducted on four small-scale, unseen datasets: PRID2011~\cite{hirzer2011classification}, GRID~\cite{loy2013manifold}, VIPeR~\cite{gray2007evaluating}, and iLIDS~\cite{li2018unsupervised}. The reported results are averaged over 10 random splits of query and gallery sets to ensure statistical reliability. In Protocol-2, a more rigorous domain generalization setting is considered. Among the four major datasets--Market-1501 (M), MSMT17 (MS), CUHK03 (C3), and CUHK-SYSU (CS)--one dataset is reserved as the target domain for testing, while the remaining three are used for training. Importantly, CUHK-SYSU (CS) is excluded from being used as a target domain in testing. This is because CS contains only a single camera view, making it unsuitable for cross-camera matching evaluation, which is a fundamental requirement in person re-identification benchmarks. Therefore, CS is used only as part of the training source domains. Protocol-2 is particularly valuable in that the training and test domains are completely disjoint, enabling the most severe form of domain shift to be evaluated. As such, it serves as a strong benchmark to assess how well a method can generalize to a completely unseen domain, which is a core objective in DG-ReID research. Protocol-3 builds upon Protocol-2 by relaxing the training constraints. While the target domain remains held out for evaluation, all available images from the remaining source domains--including both training and testing subsets--are utilized for training. This setup better reflects practical deployment scenarios, where labeled datasets from various domains are fully accessible during training. Hence, Protocol-3 is considered to be more realistic and aligned with real-world generalization requirements, while still maintaining a held-out target domain for evaluation. The reason for employing these three evaluation protocols is to comprehensively assess generalization performance under various combinations of training and test datasets. Protocol-1 evaluates ID matching performance on unseen target domains by ensuring that the test datasets are completely excluded from training. Protocol-2 tests all possible combinations of training datasets from large-scale sources, enabling evaluation of how well a method generalizes across different source domain splits. Lastly, Protocol-3 allows the use of the full training, query, and gallery sets from the source datasets, simulating a practical deployment scenario where all available source data can be leveraged. While performance generally improves with more training data, Protocol-3 assesses how well the model generalizes under these realistic conditions and provides an upper bound on expected DG-ReID performance.

\begin{sidewaystable*}
  \centering
  \footnotesize
  \caption{Performance comparisons with the state-of-the-art methods on Protocol-1. Since DukeMTMC-reID~\cite{zheng2017unlabeled} (denoted as D) has been officially withdrawn, it is excluded from our training process. The \textit{Type} column indicates the methodological category: \textit{memory} for memory-augmented methods, \textit{norm} for normalization-based methods, \textit{MoE} for mixture-of-experts-based methods, \textit{data} for data-driven learning-based methods, \textit{meta} for meta-learning-based methods, \textit{gradient} for gradient-alignment-based methods. If multiple strategies are used, they are listed using commas. Numbers in bold indicate the best performance and underscored ones are the second best.}
  \vspace{10pt}
  \renewcommand{\arraystretch}{1.5}
  \setlength{\tabcolsep}{2.5mm}
  \resizebox{\linewidth}{!}{
    \begin{tabular}{c|c|c|cc|cc|cc|cc|cc}
    \hline
    \multirow{2}{*}{Source} & \multirow{2}{*}{Method} & \multirow{2}{*}{Venue} & \multicolumn{2}{c|}{PRID} & \multicolumn{2}{c|}{GRID} & \multicolumn{2}{c|}{VIPeR} & \multicolumn{2}{c|}{iLIDs} & \multicolumn{2}{c}{Average} \\
    & &  & mAP & R1 & mAP & R1 & mAP & R1 & mAP & R1 & mAP & R1 \\
    \hline
    & DIMN~\cite{song2019dimn} & CVPR'19 & 52.0 & 39.2 & 41.1 & 29.3 & 60.1 & 51.2 & 78.4 & 70.2 & 58.0 & 47.5 \\
    \multirow{2}{*}{M+D} & SNR~\cite{jin2020style} & CVPR'20 & 66.5 & 52.1 & 47.7 & 40.2 & 61.3 & 52.9 & \underline{89.9} & \underline{84.1} & 66.3 & 57.3 \\
    \multirow{2}{*}{C2+C3+CS} & DMG-Net~\cite{bai2021person30k} & CVPR'21 & \underline{68.4} & \underline{60.6} & 56.6 & 51.0 & 60.4 & 54.0 & 84.0 & 79.3 & 67.4 & 61.3 \\
    & RaMOE~\cite{dai2021ramoe} & CVPR'21 & 67.3 & 57.7 & 54.2 & 46.8 & 64.6 & 56.6 & \textbf{90.2} & \textbf{85.0} & 69.1 & 61.5 \\
    & MDA~\cite{ni2022mda} & CVPR'22 & - & - & \textbf{62.9} & \textbf{61.2} & \textbf{71.7} & \textbf{63.5} & 84.4 & 80.4 & \underline{73.0} & \textbf{68.4} \\
    & DTIN-Net~\cite{jiao2022dtin} & ECCV'22 & \textbf{79.7} & \textbf{71.0} & \underline{60.6} & \underline{51.8} & \underline{70.7} & \underline{62.9} & 87.2 & 81.8 & \textbf{74.6} & \underline{66.9} \\
    \hline
    \multirow{6}{*}{M+C2}& M$^3$L~\cite{zhao2021memory} & CVPR'21 & 64.3 & 53.1 & 55.0 & 44.4 & 66.2 & 57.5 & 81.5 & 74.0 & 66.8 & 57.2 \\
    \multirow{6}{*}{C3+CS} & META~\cite{xu2022mimic} & ECCV'22 & 71.7 & 61.9 & 60.1 & 52.4 & 68.4 & 61.5 & 83.5 & 79.2 & 70.9 & 63.8 \\
    & MetaBIN~\cite{choi2021metabin} & CVPR'21 & \underline{81.0} & \textbf{74.2} & 57.9 & 48.4 & 68.6 & 59.9 & 87.0 & 81.3 & 73.6 & 66.0 \\
    & ACL~\cite{zhang2022adaptive} & ECCV'22 & 73.4 & 63.0 & 65.7 & 55.2 & 75.1 & 66.4 & 86.5 & 81.8 & 75.2 & 66.6 \\
    & ISR~\cite{dou2023identity} & ICCV'23 & 70.8 & 59.7 & 65.2 & 55.8 & 66.6 & 58.0 & \textbf{91.7} & \textbf{87.6} & 73.6 & 65.3 \\
    & PAOA~\cite{li2024paoa} & WACV'24 & 75.1 & 65.6 & 67.2 & 56.3 & \underline{76.6} & \underline{66.7} & 87.1 & 83.1 & 76.2 & 67.9 \\
    & BAU~\cite{cho2024alignment} & NeurIPS'24 & 77.2 & 68.4 & \textbf{68.1} & \textbf{59.8} & 74.6 & 66.1 & \underline{88.7} & \underline{83.7} & \underline{77.2} & \underline{69.5} \\
    & ReNorm~\cite{nie2025normalization} & ECCV'24 & \textbf{81.3} & \underline{73.9} & \underline{67.5} & \underline{59.4} & \textbf{77.0} & \textbf{69.1} & 87.0 & 82.2 & \textbf{78.2} & \textbf{72.2} \\
    \hline
    \end{tabular}
  }
  \label{table1}
\end{sidewaystable*}

\begin{sidewaystable*}
  \centering
  \footnotesize
  \caption{Performance comparisons with the state-of-the-art methods on Protocol-2 and 3. The \textit{Type} column indicates the methodological category: \textit{memory} for memory-augmented methods, \textit{norm} for normalization-based methods, \textit{MoE} for mixture-of-experts-based methods, \textit{data} for data-driven learning-based methods, \textit{meta} for meta-learning-based methods, and \textit{gradient} for gradient-based methods. If multiple strategies are used, they are listed using commas. Numbers in bold indicate the best performance and underscored ones are the second best.}
  \vspace{10pt}
  \renewcommand{\arraystretch}{1.5}
  \resizebox{\linewidth}{!}{
    \begin{tabular}{c|c|c|cc|cc|cc|cc}
    \hline
    \multirow{2}{*}{Setting} & \multirow{2}{*}{Method}  & \multirow{2}{*}{Venue}
      & \multicolumn{2}{c|}{M+MS+CS $\rightarrow$ C3} 
      & \multicolumn{2}{c|}{M+CS+C3 $\rightarrow$ MS} 
      & \multicolumn{2}{c|}{MS+CS+C3 $\rightarrow$ M} 
      & \multicolumn{2}{c}{Average} \\
      & & & mAP & R1 & mAP & R1 & mAP & R1 & mAP & R1 \\
    \hline
    \multirow{8}{*}{Protocol-2}
      & SNR~\cite{jin2020style} & CVPR'20 &  8.9 & 8.9 & 6.8 & 19.9 & 34.6 & 62.7 & 16.8 & 30.5 \\
      & ISR~\cite{dou2023identity} & ICCV'23 & 27.4 & 26.1 & 21.2 & 45.7 & 65.1 & 85.1 & 37.9 & 52.3 \\
      & M$^3$L~\cite{zhao2021memory} & CVPR'21 & 34.2 & 34.4 & 16.7 & 37.5 & 61.5 & 82.3 & 37.5 & 51.4 \\
      & MetaBIN~\cite{choi2021metabin} & CVPR'21 & 28.8 & 28.1 & 17.8 & 40.2 & 57.9 & 80.1 & 34.8 & 49.5 \\
      & META~\cite{xu2022mimic} & ECCV'22 & 36.3 & 35.1 & 22.5 & 49.9 & 67.5 & 86.1 & 42.1 & 57.0 \\
      & ACL~\cite{zhang2022adaptive} & ECCV'22 & 41.2 & 41.8 & 20.4 & 45.9 & \underline{74.3} & \underline{89.3} & 45.3 & 59.0 \\
      & BAU~\cite{cho2024alignment} & NIPS'24 & \underline{42.8} & \underline{43.9} & \underline{24.3} & \underline{50.9} & \textbf{77.1} & \textbf{90.4} & \textbf{48.1} & \underline{61.7} \\
      & ReNorm~\cite{nie2025normalization} & ECCV'24 & \textbf{43.6} & \textbf{44.7} & \textbf{25.6} & \textbf{55.6} & 72.7 & 89.1 & \underline{47.3} & \textbf{63.1} \\
    \hline
    \multirow{8}{*}{Protocol-3}
      & SNR~\cite{jin2020style} & CVPR'20 & 17.5 & 17.1 & 7.7 & 22.0 & 52.4 & 77.8 & 25.9 & 39.0 \\
      & M$^3$L~\cite{zhao2021memory} & CVPR'21 & 35.7 & 36.5 & 17.4 & 38.6 & 62.4 & 82.7 & 38.5 & 52.6 \\
      & MetaBIN~\cite{choi2021metabin} & CVPR'21 & 43.0 & 43.1 & 18.8 & 41.2 & 67.2 & 84.5 & 43.0 & 56.3 \\
      & META~\cite{xu2022mimic} & ECCV'22 & 47.1 & 46.2 & 24.4 & 52.1 & 76.5 & 90.5 & 49.3 & 62.9 \\
      & ACL~\cite{zhang2022adaptive} & ECCV'22 & 49.4 & 50.1 & 21.7 & 47.3 & 76.8 & 90.6 & 49.3 & 62.7 \\
      & PAOA~\cite{li2024paoa} & WACV'24 & 49.8 & 50.5 & 25.1 & 51.5 & 77.1 & 90.8 & 50.7 & 64.3 \\
      & BAU~\cite{cho2024alignment} & NeurIPS'24 & \underline{50.6} & \underline{51.8} & \underline{26.8} & \underline{54.3} & \textbf{79.5} & \underline{91.1} & \underline{52.3} &  \underline{65.7} \\
      & ReNorm~\cite{nie2025normalization} & ECCV'24 & \textbf{51.9} & \textbf{52.5} & \textbf{27.8} & \textbf{57.8} & \underline{77.4} & \textbf{91.2} & \textbf{52.4} & \textbf{67.2} \\
    \hline
    \end{tabular}
  }
  \label{table2}
\end{sidewaystable*}

\subsection{Performance Comparison of State-of-the-Art Methods}

\begin{figure}[t!]
  \centering
  \begin{tikzpicture}
    \node[inner sep=0pt] (img) at (0,0) {\includegraphics[width=\linewidth]{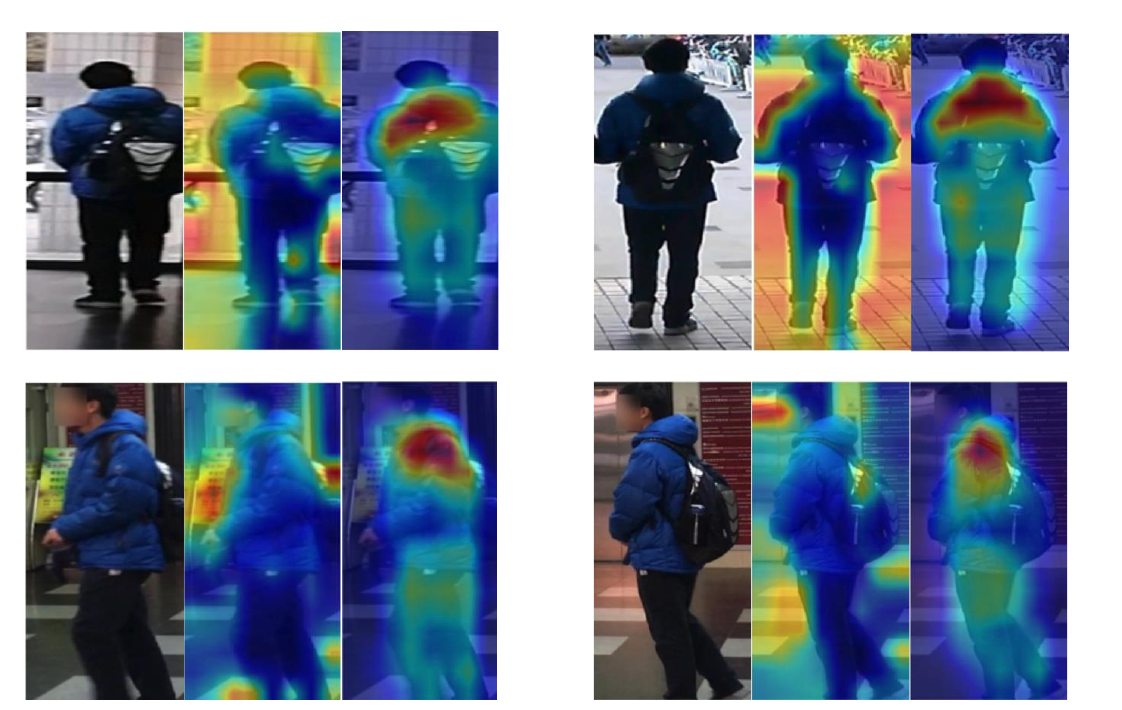}};
    \node[anchor=north west] at ([xshift=25pt,yshift=-250pt]img.north west) {\footnotesize(a)};
    \node[anchor=north east] at ([xshift=-286pt,yshift=-250pt]img.north east) {\footnotesize(b)};
    \node[anchor=north east] at ([xshift=-235pt,yshift=-250pt]img.north east) {\footnotesize(c)};
    \node[anchor=north east] at ([xshift=-145pt,yshift=-250pt]img.north east) {\footnotesize(d)};
    \node[anchor=north east] at ([xshift=-90pt,yshift=-250pt]img.north east) {\footnotesize(e)};
    \node[anchor=north east] at ([xshift=-35pt,yshift=-250pt]img.north east) {\footnotesize(f)};
  \end{tikzpicture}
  \vspace{-20pt}
  \caption{Attention map visualization of embedding vectors on MSMT17 query images. (a), (d): original images; (b), (e): attention maps from MetaBIN~\cite{choi2021metabin}; (c), (f): attention maps from ACL~\cite{zhang2022adaptive}. All images belong to the same identity. The models highlight different body regions depending on their learned feature importance.}
  \vspace{-10pt}
  \label{fig:att}
\end{figure}

Following prior studies~\cite{jin2020style, zhang2022adaptive, xu2022mimic, nie2025normalization, choi2021metabin, dai2021ramoe, zhao2021memory, cho2024alignment, chang2019dsbn}, most existing works adopt ResNet-50~\cite{he2016deep} pre-trained on ImageNet, often with instance normalization layers, as the backbone network. All input images are typically resized to 256$\times$128. For a fair comparison, the same training protocol is employed by SNR~\cite{jin2020style}, RaMoE~\cite{dai2021ramoe}, MDA~\cite{chang2019dsbn}, M$^3$L~\cite{zhao2021memory}, META~\cite{xu2022mimic}, MetaBIN~\cite{choi2021metabin}, ACL~\cite{zhang2022adaptive}, ISR~\cite{dou2023identity}, and ReNorm~\cite{nie2025normalization}, using a batch size of 64 composed of 16 identities with 4 images each. BAU~\cite{cho2024alignment} adopts a similar setting, sampling 256 images per iteration with 64 identities and 4 instances per identity. Standard data augmentations, including random flipping, cropping, random erasing~\cite{zhong2020random}, RandAugment~\cite{cubuk2020randaugment}, and color jittering, are commonly applied.

\paragraph{Protocol-1 Results}  
Performance comparisons under Protocol-1 are summarized in Table~\ref{table1}. In this setting, DukeMTMC-reID (denoted as D)~\cite{zheng2017unlabeled}, which has been officially withdrawn, is excluded from the training datasets. As shown in Table~\ref{table1}, despite using fewer source datasets, some of the more recent methods (listed in the lower part) outperform earlier approaches in terms of average mAP and Rank-1 across the four test sets. Notably, ReNorm~\cite{nie2025normalization} achieves the best average mAP and Rank-1 accuracy, outperforming the second-best method, BAU~\cite{cho2024alignment}, by 1.0\% and 2.7\%, respectively.

\paragraph{Protocol-2 Results}  
Further comparisons under Protocol-2 is presented in the upper part of Table~\ref{table2}. In this setting, only the training images of each source domain are used, which generally results in lower performance compared to Protocol-3. DIMN~\cite{song2019dimn}, DMG-Net~\cite{bai2021person30k}, M$^3$L~\cite{zhao2021memory}, RaMoE~\cite{dai2021ramoe} DTIN-Net~\cite{jiao2022dtin} and MDA~\cite{chang2019dsbn} are excluded from this comparison due to no results reported for Protocol-2 and -3. While the average mAP of BAU~\cite{cho2024alignment} is 0.4\% higher, ReNorm~\cite{nie2025normalization} achieves a 1.4\% improvement in Rank-1 accuracy. Although many existing methods rely on normalization-based~\cite{jin2020style, zhang2022adaptive, xu2022mimic, nie2025normalization, choi2021metabin} or meta-learning-based approaches~\cite{zhang2022adaptive, dai2021ramoe}, it is also noteworthy that BAU achieves these results without employing complex normalization techniques or domain-specific network architectures. Overall, both BAU~\cite{cho2024alignment} and ReNorm~\cite{nie2025normalization} demonstrate consistently strong performance across the reported benchmarks.

\paragraph{Protocol-3 Results}  
Protocol-3 utilizes both the training and testing images from each source domain, generally leading to improved overall performance, while maintaining the same relative ranking trends as Protocol-2. In the lower section of Table~\ref{table2}, ReNorm~\cite{nie2025normalization} reports the highest average mAP and Rank-1 accuracy among all evaluated methods. While recent DG-ReID methods are approaching performance saturation, the relatively low mAP of 27.8\% on the target domain MS under Protocol-3 indicates potential for further improvement. 

To further investigate why MSMT17 remains particularly challenging under Protocol-3, we analyze the attention behavior of two representative DG-ReID models, MetaBIN~\cite{choi2021metabin} and ACL~\cite{zhang2022adaptive}, using attention maps shown in Fig.~\ref{fig:att}. Both models share a ResNet-50 backbone for fair comparison. The attention maps are generated by applying the learned classification weights to the final convolutional feature maps before the fully connected (FC) layer. We visualize the attention of each model for the same identity appearing in different scenes. As illustrated in Fig.~\ref{fig:att} (b), (e), MetaBIN often attends excessively to background regions, occasionally more than to the target person, indicating susceptibility to domain-specific noise. In contrast, ACL (Fig.~\ref{fig:att} (c), (f)) exhibits more stable person-centric attention, relatively unaffected by background clutter. This qualitative difference aligns with the mAP results on MSMT17 under Protocol-3, where ACL outperforms MetaBIN (21.7\% vs. 18.8\%). However, in ACL, the conservative focus may neglect identity-discriminative details such as backpacks or accessories, potentially limiting its discriminative capacity. These findings suggest that suboptimal attention allocation--either overly diffuse or overly narrow--may hinder both models from fully capturing robust identity cues, thereby contributing to the overall performance degradation observed on challenging domains like MSMT17.

\section{Future Research Direction}
\label{sec6}
We present insights into future research directions for Domain Generalizable Person Re-identification (DG-ReID), focusing on video-based, camera-aware and frequency-based methods. These areas are still largely unexplored in DG-ReID and present opportunities for significant advancements in the field. Additionally, extending DG-ReID to Domain Generalizable Person Search (DG-Person Search) introduces further challenges, requiring consideration of both person localization and re-identification within raw images.

\subsection{Video-based DG-ReID}
\label{subsec6-1}
Although video datasets for person re-identification (ReID)~\cite{zheng2016mars, wu2018oneshot, li2018tracklet} have been widely studied in the general ReID community, domain generalization in video-based ReID settings remains largely unexplored. This is in stark contrast to real-world surveillance systems, which primarily rely on continuous video streams rather than isolated still images. The lack of DG-ReID research in video settings presents a critical gap between academic progress and practical deployment. Video-based datasets inherently provide richer information, capturing both spatial and temporal dynamics across consecutive frames. These include motion cues, viewpoint transitions, and pose variations over time, which are often essential for robust identity matching. For instance, gait and body movement patterns provide temporal cues that are less sensitive to visual appearance changes and may carry more domain-invariant characteristics. A promising direction is to explicitly learn temporal identity-invariance across videos. For example, CION~\cite{zuo2024cion} proposes a cross-video contrastive learning strategy that correlates images of the same person across different videos. Unlike traditional instance- or tracklet-level methods that focus on augmentations within a single video, CION introduces a learning paradigm that aligns positive pairs from different video sequences to capture cross-video identity consistency. While CION is not originally designed for domain generalization, its core strategy--cross-video identity correlation--can be naturally extended to DG-ReID settings. One possible direction is to integrate a domain-aware projection head (\textit{e.g.}, MLP) that maps features into domain-specific subspaces while maintaining identity consistency across domains. By learning contrastive objectives within each domain while aligning features across domains, such an approach encourages domain-aware yet generalizable representation learning. Alternatively, one could adopt 3D convolutional backbones such as I3D~\cite{carreira2017i3d} or SlowFast~\cite{feichtenhofer2019slowfast} to extract temporal motion features, then disentangle them from appearance features. A domain classifier with gradient reversal can then be applied to the motion component to enforce domain confusion, promoting domain-invariant motion encoding. This strategy explicitly regularizes frame-level temporal patterns to avoid overfitting to domain-specific styles such as background or lighting dynamics. By treating each source video domain independently and mining inter-domain temporal correspondences, one can encourage the model to learn identity-invariant features that are robust to both visual and contextual domain shifts. Therefore, integrating such cross-video alignment into DG frameworks opens a promising direction for future research. To fully enable this line of work, the development of large-scale, multi-domain video benchmarks and efficient video-based training pipelines is also needed.

\subsection{Camera-aware DG-ReID}
\label{subsec6-2}
ReID datasets typically provide both person ID and camera ID labels for each image. However, despite the growing interest in domain generalization for person re-identification (DG-ReID), camera ID information remains an underutilized resource. Most DG-ReID approaches focus on dataset-level domain shifts, often neglecting camera-specific variations that are inherently embedded within each dataset. These variations--such as differences in viewpoint, resolution, illumination, and occlusion--are significant contributors to intra- and inter-domain discrepancies. Camera IDs, by design, encapsulate such domain-specific characteristics~\cite{zheng2015scalable}, and have proven useful in conventional ReID settings. Several works have effectively leveraged this information; for instance, Li et al.~\cite{li2022cameraaware} proposed camera-aware adaptive learning, and CaCL~\cite{lee2023cameradriven} utilized a curriculum learning strategy based on camera-wise clustering to gradually align representations across cameras and domains. Extending camera-aware learning strategies to DG-ReID holds great potential. One promising direction is to adopt camera-adaptive normalization, where the mixing ratio between BN and IN is not fixed or globally learnable, but instead conditioned on the camera ID. Unlike previous normalization-based methods~\cite{choi2021metabin, jin2020style} that statically assign or learn a global IN/BN ratio, this approach dynamically adjusts the normalization behavior based on the camera label, allowing the model to directly mitigate camera-specific biases during feature normalization. Another potential approach involves camera-aware prompting in transformer-based architectures. By using a Vision Transformer (ViT)~\cite{dosovitskiy2020vit}, one can insert a camera-specific token or prompt embedding into the input sequence to guide the attention mechanism toward camera-aware feature extraction. Compared to convolutional backbones like ResNet~\cite{he2016deep}, which emphasize local structures, transformers can capture global dependencies, making them well-suited for disentangling subtle camera-related cues. Moreover, this design can be integrated with prompt tuning-based DG modules~\cite{zhao2025clipfgdi}, further enhancing adaptability to camera variation during both training and inference. These approaches suggest that better leveraging camera ID information--whether through normalization adaptation or attention guidance--offers a promising yet underexplored direction for improving generalization in DG-ReID.

\begin{figure}[t!]
    \centering
    \includegraphics[width=\linewidth]{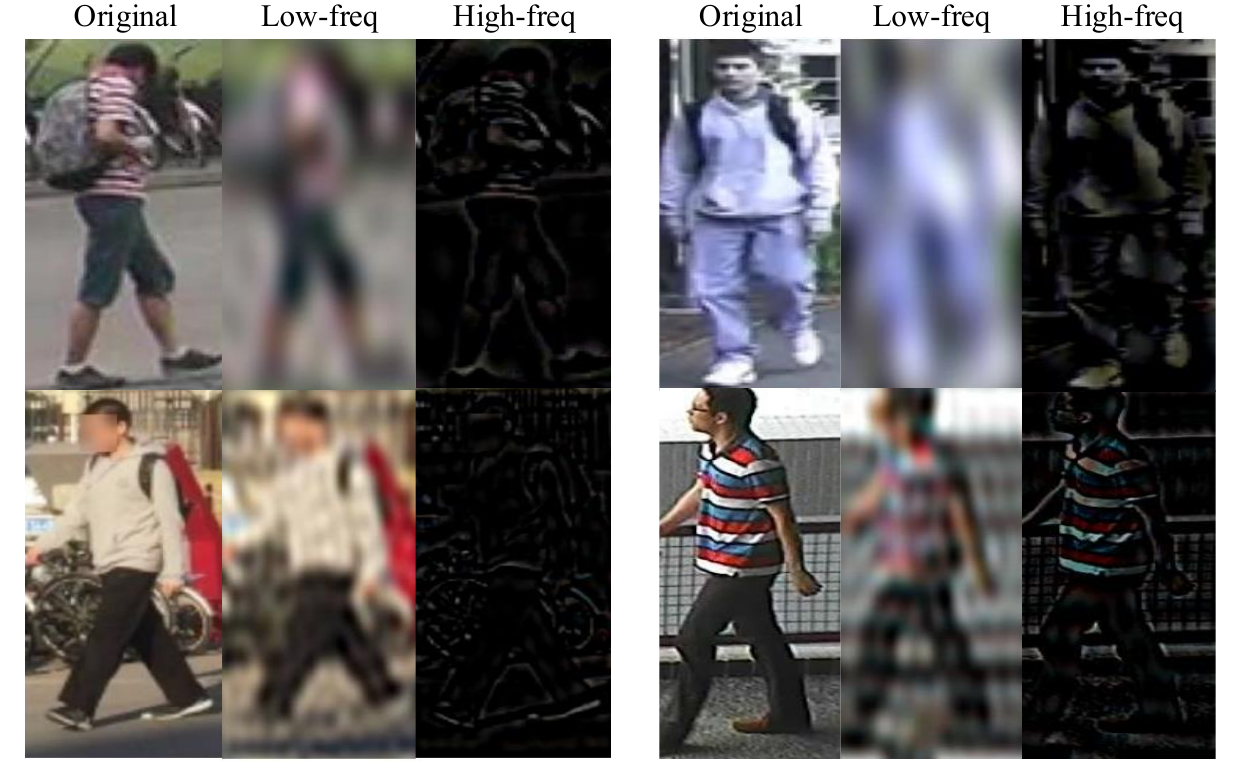}
    \vspace{-10pt}
    \caption{Visualization of frequency decomposition results using samples from four different ReID datasets. Each triplet shows the original image (left), low-frequency reconstruction (center), and high-frequency reconstruction (right). High-frequency components capture fine details such as textures and edges, while low-frequency components preserve coarse structural and color information. This highlights how frequency decomposition separates complementary cues for ReID.}
    \vspace{-10pt}
    \label{freq}
\end{figure}

\subsection{Frequency-based DG-ReID}
\label{subsec6-3}
Frequency-domain methods have recently gained attention in domain generalization research due to their ability to extract complementary information beyond the spatial domain~\cite{wang2023frequencyshortcut, wang2022frequencydisentanglement, he2024combating}. These methods leverage spectral representations, particularly high-frequency components, to uncover subtle and discriminative signals that are often suppressed or ignored by spatial convolutions. For instance, high-frequency information often encodes fine-grained edge structures and detailed textures that are less sensitive to domain-specific styles. Specifically, FFDI~\cite{wang2022frequencydisentanglement} proposes a frequency-based disentanglement framework that explicitly separates domain-invariant and domain-specific components in the spectral domain. By isolating frequency bands associated with biased features, the method reduces domain overfitting and improves cross-domain robustness. This approach highlights the unique strength of frequency analysis in capturing latent semantic cues that are not easily separable in the spatial domain. As visualized in Fig.~\hyperref[freq]{9}, frequency decomposition enables a clear separation between low-frequency information, which mainly reflects illumination and color bias, and high-frequency components that preserve fine structural details such as edges and clothing contours. Notably, we observe that identity-relevant cues such as striped clothing patterns are often retained in the high-frequency bands, while overall structural and color information remains in the low-frequency bands. The visual quality of these reconstructions is also positively correlated with the resolution and brightness of the original image, highlighting the importance of frequency fidelity in domain-invariant representation learning. This separation lays the foundation for frequency-aware learning strategies that aim to decouple domain-specific styles from identity-relevant structures, and has inspired a series of methods leveraging frequency-domain transformations to improve generalization under distribution shifts. Building upon this insight, one promising strategy is to construct a dual-stream network that separately processes low- and high-frequency components. For instance, using Discrete Wavelet Transform (DWT) or FFT, the image is decomposed into sub-bands such as LL (low-low), LH (low-high), HL (high-low), and HH (high-high). The LL band, containing low-frequency information, is fed into a stream responsible for capturing domain-specific styles (\textit{e.g.}, color tone or lighting), while the remaining high-frequency bands are processed by a stream that focuses on preserving identity-relevant structures. To enforce disentanglement between the two streams, losses such as orthogonality constraints or mutual information minimization are employed, thereby ensuring minimal feature interference and promoting domain-invariant representation learning. Another line of work explores frequency disentanglement from the perspective of amplitude and phase. Since amplitude reflects the magnitude of frequency components and phase encodes structural alignment, it is possible to mix the amplitude from different domains while keeping the phase fixed to preserve identity structure. This allows the model to simulate domain shifts in the frequency domain while maintaining identity consistency. A phase-preserving loss (\textit{e.g.}, L2 distance between phases) can be applied to retain structural information. This approach treats the mixed input as an implicit domain augmentation, improving robustness to distribution shifts without relying on spatial transformations. In the context of person re-identification (ReID), which is particularly vulnerable to appearance-based domain shifts, frequency-based augmentation and decomposition have shown promising results. For example, PHA~\cite{zhang2023pha} demonstrates that manipulating frequency content in the input images leads to more robust representations under distribution shifts. This opens a new perspective for DG-ReID, where frequency-domain modeling can be integrated into existing pipelines—either through spectral augmentation, frequency-aware loss functions, or dual-domain feature fusion. Moreover, frequency-domain techniques may provide a more principled way to counteract camera-specific distortions, such as blur, compression artifacts, and lighting inconsistencies, which often manifest in specific frequency ranges. By training models to be invariant to such perturbations, it is possible to build ReID systems that generalize better across both seen and unseen domains. Overall, the frequency domain offers a rich and largely underexplored avenue for advancing DG-ReID.

\subsection{DG-Person Search}
\label{subsec6-4}
Person Search~\cite{zheng2017wild, xiao2017joint, li2021sequential} extends the traditional ReID task by requiring the model to simultaneously localize and identify individuals in uncropped raw images. Unlike conventional ReID, where person bounding boxes are pre-annotated, Person Search reflects more realistic surveillance scenarios in which person detection and re-identification must be conducted jointly. This integration introduces additional complexity, as the system must handle variations not only in appearance but also in detection quality, background clutter, and occlusions. Despite its practical relevance, robust generalization across domains in Person Search remains a largely unexplored area. Most existing Person Search models are trained and evaluated within the same dataset or under fixed conditions, leaving their cross-domain robustness under question. Domain Generalizable Person Search (DG-Person Search) emerges as a natural and necessary extension of DG-ReID, addressing the combined challenges of localization and identification under domain shift. Pursuing DG-Person Search poses unique challenges that go beyond those in DG-ReID. First, domain-specific variations can affect detection (\textit{e.g.}, different camera angles, resolutions, and lighting) and re-identification (\textit{e.g.}, clothing styles or occlusions) in different ways, requiring tailored strategies for each subtask. Second, these two tasks may exhibit conflicting optimization objectives, making it difficult to jointly generalize across domains without sacrificing one aspect of performance. This creates a multi-task generalization problem that has yet to be properly addressed in existing literature. To mitigate this, one promising direction is to decompose the task using a domain-specific detector and a domain-invariant ID encoder. In this setting, the detection module is fine-tuned independently for each source domain to capture domain-specific scene characteristics, such as camera angles or background clutter by training a Faster R-CNN~\cite{ren2015faster} per domain or by adapting backbone layers without freezing. Once person bounding boxes are obtained, the cropped images are fed into a shared ReID encoder, which is meta-learned to generalize across domains. This enables robust identity representation learning regardless of the detection domain and helps bridge domain gaps in the re-identification stage. An alternative solution is to utilize a scene-conditioned person search approach, in which background scene information serves as a proxy for domain cues. Since scenes tend to be consistent within a domain (\textit{e.g.}, mall vs. airport), they offer a natural signal for learning domain-aware representations. One way to implement this is to extract background descriptions using vision-language models such as CLIP~\cite{radford2021clip}, focusing on the non-person regions of the image. The resulting scene captions or embeddings can then be used to compute an auxiliary alignment loss, grouping samples with similar scene descriptors as part of the same domain. This method is particularly valuable when explicit camera or domain labels are unavailable, as the scene embedding provides a soft, data-driven domain signal. These strategies highlight the potential of DG-Person Search to extend beyond traditional paradigms. By integrating domain-specific detection with domain-invariant identification or leveraging scene cues as domain proxies, future methods can better address the multifaceted nature of cross-domain generalization. Ultimately, advancing research in DG-Person Search could significantly benefit real-world applications such as cross-site surveillance, drone-based tracking, and public safety monitoring, where system deployment in unseen environments is inevitable. Thus, exploring domain generalization in the context of Person Search not only bridges a critical research gap but also expands the applicability of DG-ReID techniques to more comprehensive and practical settings.

\section{Conclusion}
\label{sec7}
This paper has presented a comprehensive survey of Domain Generalizable Person Re-identification (DG-ReID). We have reviewed the standard pipeline underlying DG-ReID methods and systematically categorized existing approaches into seven modules: normalization-based, mixture-of-experts-based, memory-based, meta-learning-based, data-driven, CLIP-based, and others. We have also conducted a case study on VI-ReID, a representative task that shares the core challenge of domain shift, to provide insights into broader applicability. Additionally, we have summarized commonly used datasets and evaluation protocols, and compared the performance of state-of-the-art methods across standard benchmarks. While DG-ReID has achieved notable progress, it still faces several unresolved challenges. To this end, we have outlined promising research directions, particularly in video-based, camera-aware, frequency-based, and DG-person search approaches. Finally, we encourage extending the scope of DG-ReID toward person search and other real-world applications, thereby facilitating broader and more practical advancements in the field.

\section{Acknowledgments}
This work was supported by both the MSIT(Ministry of Science and ICT), Korea, under the Graduate School of Metaverse Convergence support program(IITP-2024-RS-2024-00418847) supervised by the IITP(Institute for Information \& Communications Technology Planning \& Evaluation, and the Chung-Ang University Research Scholarship Grants in 2023.

\bibliographystyle{elsarticle-num} 
\bibliography{reference}     
\end{document}